\documentclass[sigconf,screen]{acmart}
\copyrightyear{2023}
\acmYear{2023}
\setcopyright{acmlicensed}\acmConference[MM '23]{Proceedings of the 31st
ACM International Conference on Multimedia}{October 29-November 3,
2023}{Ottawa, ON, Canada}
\acmBooktitle{Proceedings of the 31st ACM International Conference on
Multimedia (MM '23), October 29-November 3, 2023, Ottawa, ON, Canada}
\acmPrice{15.00}
\acmDOI{10.1145/3581783.3611808}
\acmISBN{979-8-4007-0108-5/23/10}

\AtBeginDocument{%
  \providecommand\BibTeX{{%
    Bib\TeX}}}

\usepackage{mathtools}
\usepackage[vlined,ruled]{algorithm2e}
\usepackage{algpseudocode}
\usepackage{subfig}
\usepackage{multirow}
\usepackage{booktabs}
\usepackage{pifont}
\usepackage{bm}
\usepackage{enumitem}
\usepackage{xr}
\makeatletter

\newcommand{\eat}[1]{}

\AtBeginDocument{%
  \providecommand\BibTeX{{%
    \normalfont B\kern-0.5em{\scshape i\kern-0.25em b}\kern-0.8em\TeX}}}

\begin{document}

\begin{sloppypar}
\title{Cal-SFDA: Source-Free Domain-adaptive Semantic Segmentation with Differentiable Expected Calibration Error}

\author{Zixin Wang}
\affiliation{%
  \institution{The University of Queensland}
  \country{Australia}
}
\email{zixin.wang@uq.edu.au}

\author{Yadan Luo}
\affiliation{%
  \institution{The University of Queensland}
  \country{Australia}
}
\email{y.luo@uq.edu.au}

\author{Zhi Chen}
\affiliation{%
  \institution{The University of Queensland}
  \country{Australia}
}
\email{zhi.chen@uq.edu.au}

\author{Sen Wang}
\affiliation{%
  \institution{The University of Queensland}
  \country{Australia}
}
\email{sen.wang@uq.edu.au}

\author{Zi Huang}
\affiliation{%
  \institution{The University of Queensland}
  \country{Australia}
}
\email{helen.huang@uq.edu.au}

\begin{abstract}
The prevalence of domain adaptive semantic segmentation has prompted concerns regarding source domain data leakage, where private information from the source domain could inadvertently be exposed in the target domain. To circumvent the requirement for source data, source-free domain adaptation has emerged as a viable solution that leverages self-training methods to pseudo-label high-confidence regions and adapt the model to the target data. However, the confidence scores obtained are often highly \textit{biased} due to overconfidence and class-imbalance issues, which render both model selection and optimization problematic. In this paper, we propose a novel \underline{cal}ibration-guided \underline{s}ource-\underline{f}ree \underline{d}omain \underline{a}daptive semantic segmentation (\textbf{Cal-SFDA}) framework. The core idea is to estimate the expected calibration error (ECE) from the segmentation predictions, serving as a strong indicator of the model's generalization capability to the unlabeled target domain. The estimated ECE scores, in turn, assist the model training and fair selection in both source training and target adaptation stages. During model pre-training on the source domain, we ensure the differentiability of the ECE objective by leveraging the LogSumExp trick and using ECE scores to select the best source checkpoints for adaptation. To enable ECE estimation on the target domain without requiring labels, we train a value net for ECE estimation and apply statistic warm-up on its BatchNorm layers for stability. The estimated ECE scores assist in determining the reliability of prediction and enable class-balanced pseudo-labeling by positively guiding the adaptation progress and inhibiting potential error accumulation. Extensive experiments on two widely-used synthetic-to-real transfer tasks show that the proposed approach surpasses previous state-of-the-art by up to $5.25\%$ of mIoU with fair model selection criteria. 
\end{abstract}


\vspace{-1ex}
\maketitle
\section{Introduction}
\begin{figure}[t]\vspace{-1ex}
\centering
\includegraphics[width=0.93\linewidth]{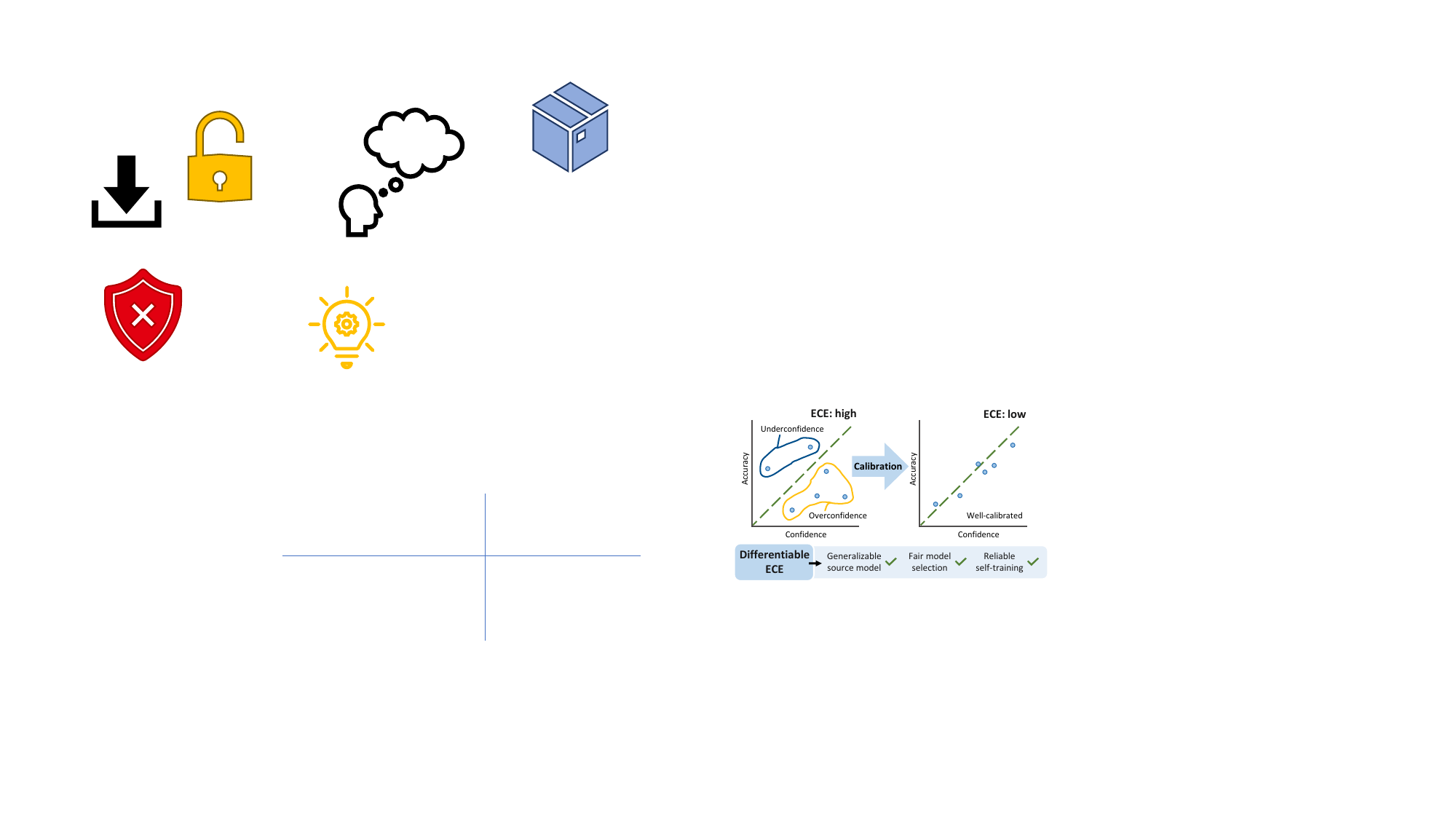}\vspace{-1ex}
\caption{The main motivation for the proposed Cal-SFDA. Minimizing differentiable ECE facilitates model optimization and selection in SFDA. \vspace{-3ex}}\label{fig:idea}
\vspace{-0.32cm}
\end{figure}
Domain adaptation (DA) \cite{DBLP:journals/ijon/WangD18,DBLP:journals/pami/KouwL21, DBLP:conf/mm/WangFCYHY18,DBLP:conf/icmcs/WangLZ0H22,DBLP:conf/nips/DongFLSL21} has shown exceptional efficacy in addressing distribution shifts \cite{DBLP:conf/mm/ChenL0QLH21} in semantic segmentation \cite{dong2021and} datasets, where the target data differs from the training data due to different locations, lighting, or weather conditions. Nevertheless, mainstream DA solutions necessitate the presence of both source and target data during model adaptation, which constrains its applicability in scenarios where these data cannot coexist. To remove the dependence on the labeled source data, source-free domain adaptive (SFDA) semantic segmentation has been developed. Current approaches for target adaptation involve generating fake source data \cite{DBLP:conf/cvpr/LiuZW21,DBLP:conf/mm/Ye0OY21} or creating multi-source environments \cite{DBLP:conf/iccv/KunduKSJB21} to improve the generalization of the source model while maintaining the privacy of the source domain. However, this approach leads to significant hardware overhead. An alternative method decomposes the training process into two by sharing source pre-trained weights into the target domain and performing \textit{confidence-based self-training}. 
In the context of confidence-based self-training, there are significant risks associated with relying solely on model \textit{confidence} for decision-making.  
Nevertheless, a substantial concern emerges in the shape of over- or under-confidence, as described in \cite{DBLP:conf/nips/KarandikarCTLSM21}, whereby a segmentation model, when trained on the source data over a long-time optimization, may lead to the overfitting of the model's parameters to the style of the source domain.
As a result, the model's ability to recognize potential new styles in target can be severely impeded, significantly compromising the source model's generalization ability.

The biased confidence not only reduces the generalization capabilities of the source model but also presents significant obstacles to target adaptation. In conventional pseudo-labeling, high-confidence pixels are assigned predicted labels regardless of correctness. As the adaptation of the target data heavily relies on these labels, this can have a cumulative effect on the model's adaptability. For instance, a tail class like `train' may be wrongly predicted and then labeled with high confidence as an incorrect class. Conversely, head classes like `road' often exhibit overconfidence, leading to a significant bias in pseudo-labeling head and tail classes and consequently hindering the model's adaptation. Accordingly, a question arises: \textit{Can a more reliable indicator of model generalization and adaptation be proposed to replace the biased confidences?}

\textbf{Model calibration} \cite{DBLP:conf/icml/GuoPSW17, DBLP:journals/corr/abs-2106-09613} is one of the natural solutions to address the biased confidence issue by adjusting the output probabilities better to reflect the true likelihood of the predicted classes. The expected calibration error (\textbf{ECE}) score, as introduced in \cite{DBLP:conf/icml/GuoPSW17}, is a commonly employed measure of model calibration that quantifies the disparity between the accuracy and the corresponding predicted confidence. A lower value of ECE indicates a higher degree of model calibration, signifying the potential for greater model generalizability. Moreover, as \cite{DBLP:conf/nips/MindererDRHZHTL21} mentioned, a lower ECE score may also refers to a better out-of-distribution (OOD) performance. 

In this work, we propose a calibration-guided source-free domain adaptive semantic segmentation framework (Cal-SFDA), which aims to fully take advantage of ECE scores to achieve (1) generalizable source training, (2) fair model selection and (3) reliable pseudo-labeling for target adaptation, as shown in Figure \ref{fig:idea}. To be specific, we first train a generalizable source segmentation model by optimizing the ECE objectives. However, one severe issue is that this ECE is \textit{non-differentiable} and thus cannot be optimized directly. To overcome this, our solution is to \textbf{differentiate ECE} by replacing the non-differentiable $\text{max}(\cdot)$ operation with its approximation $\text{LogSumExp}(\cdot)$ function. Then we jointly optimize this differentiable ECE with segmentation loss to train a generalizable source model. Furthermore, the ECE score is applied to fairly select the best source checkpoint for adaptation without the target data. 

Due to the absence of target annotations, a direct calibration evaluation on the target domain is unfeasible. To obtain the ECE score on the target domain, we build an extra branch on top of the segmentation feature extractor, called value net, to predict the ECE score per given source image. Then the value net is optimized by matching the estimated ECE and the ground-truth ECE. Subsequently, this estimated ECE is leveraged as a guiding principle during the third stage to assist in selecting the confidence threshold of each class. This could potentially disregard the ``unreliable data'' to avoid error accumulation. For example, samples with high confidence that are predicted as incorrect classes could be filtered out. During adaptation, we first use a statistic warm-up to stabilize the learning process. In order to alleviate the discrepancy between the source and target domains, self-training is then conducted based on the ECE-guided confidence threshold. Through experimental assessment, we demonstrate the effectiveness of our proposed model in improving both semantic segmentation performance and model calibration. To sum up, our \textbf{contributions} are as follows\footnote{Our code is available at \url{https://github.com/Jo-wang/Cal-SFDA}.}:
\vspace{-1ex}
\begin{itemize}[leftmargin=12pt]
    \item Our research suggests a novel approach called Cal-SFDA, which explores ECE scores as an effective generalization indicator for both model optimization and selection.
    \item We propose a simple yet effective LogSumExp strategy to differentiate ECE, thereby facilitating the source model's training. 
    \item The estimated ECE scores help eliminate erroneous pseudo-labels with high confidence during the target adaptation phase, thus mitigating the error accumulation problem.
    \item A thorough analysis is conducted on the model selection strategies during the source pre-training and target adaptation stages.
    \item Extensive experiments on two widely-used semantic segmentation transfer tasks demonstrate improvements of up to $5.25\%$ in terms of mIoU, affirming the efficacy of our approach.
\end{itemize}
\section{Related Work}
\subsection{SFDA Semantic Segmentation}
Within the realm of extensive deep vision tasks \cite{DBLP:conf/iccv/ChenLQ0H0021,DBLP:conf/mm/Chen00H20}, there has been a noticeable surge of interest in source-free domain adaptive semantic segmentation \cite{DBLP:conf/eccv/ZouYKW18,DBLP:conf/cvpr/VuJBCP19,DBLP:conf/cvpr/0001S20} in recent years. It aims to adapt a semantic segmentation model trained on a source domain to perform well on a target domain without access to the source data and target annotations. To achieve this goal, mainstream works attempt to recover the source domain knowledge by synthesizing fake samples \cite{DBLP:conf/cvpr/LiuZW21} or training a generalizable source model using multiple augmentations with multiple classification heads \cite{DBLP:conf/iccv/KunduKSJB21,DBLP:conf/icml/KunduKBMKJR22}. Negative learning \cite{DBLP:conf/iccv/KimYYK19} is also employed in self-training processes \cite{DBLP:conf/mm/YouLZCH21} to force the model to learn which category a pixel does not belong to. \cite{DBLP:journals/corr/abs-2107-10140} leverages the shared architecture across different views of the target data and proposes self-supervised selective self-training. In addition, class balancing is applied to deal with the long-tailed data distribution problem \cite{DBLP:journals/corr/abs-2110-04596} that exists broadly in image segmentation tasks \cite{DBLP:conf/icmcs/YangKH22}. More challenging settings \cite{DBLP:conf/nips/HuangGXL21,DBLP:journals/tcsv/ZhaoZLLS22} are tackled under the source-free setting. Unlike existing work, this paper focuses on achieving source-free adaptation by calibrating the model on both the source and target sides.
\vspace{-1ex}
\begin{figure*}[ht]\vspace{-3ex}
\centering
\includegraphics[width=1\linewidth]{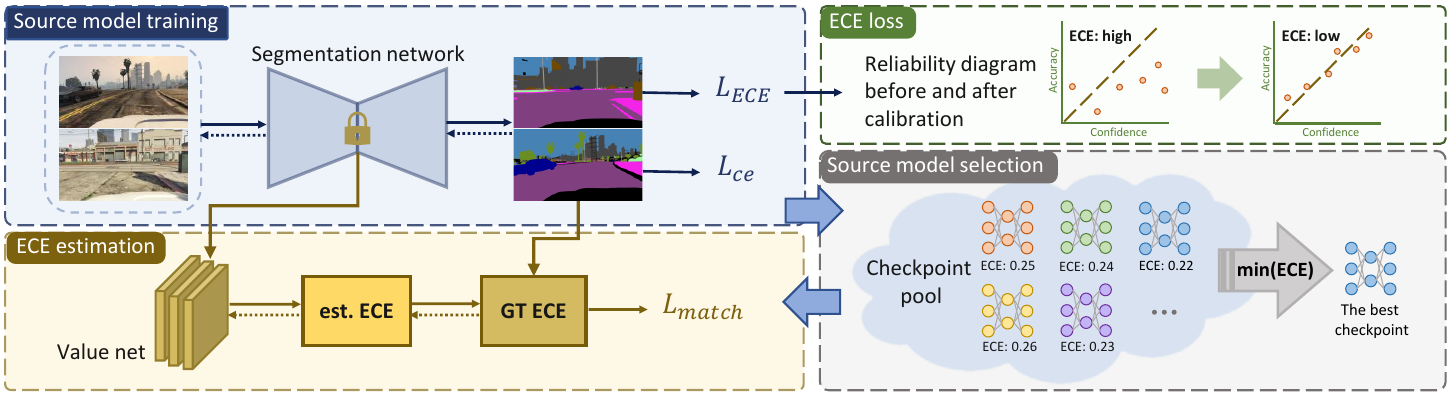}\vspace{-3ex}
\caption{An illustration of model training workflow with the source data. The differentiable ECE is proposed to lead to a better generalization performance on the unseen target domain. The value net is trained to produce estimated ECE scores and match the ground-truth ECE. The selection of the optimal source model is based on the dataset-level ECE score.\vspace{-2ex}}\label{fig:src}
\end{figure*}

\subsection{Model Calibration}
The widespread use of deep learning models in supporting real-world decision-making necessitates the calibration of predicted results to improve their reliability. A line of methods \cite{DBLP:conf/aaai/NaeiniCH15,degroot1983comparison,DBLP:conf/aistats/ParkBWL20} has been proposed to qualitatively and quantitatively measure calibration. Expected calibration error (ECE) is computed by dividing the output confidence into bins and averaging the absolute difference between accuracy and corresponding predicted confidence in each bin. The reliability diagram \cite{degroot1983comparison} is a graphical representation that visually measures the calibration of a model. A well-calibrated model will have a reliability diagram that closely follows the diagonal line, indicating that the model's predictions are accurate and reliable. Common calibration techniques, e.g.,  Histogram binning \cite{DBLP:conf/icml/ZadroznyE01} and Platt scaling \cite{platt1999probabilistic}, calibrates the model in a parametric way. While temperature scaling \cite{DBLP:conf/icml/GuoPSW17} and scaling-binning calibrator \cite{DBLP:conf/nips/KumarLM19} is more efficient and accurate. Dirichlet calibration \cite{DBLP:journals/corr/abs-1910-12656} and the focal loss \cite{DBLP:conf/nips/MukhotiKSGTD20} are additional techniques that could improve model calibration. Model calibration is also considered in the case of semantic segmentation pseudo-labeling \cite{DBLP:conf/iclr/ZouZZLBHP21} and OOD-generalization \cite{DBLP:conf/nips/WaldFGS21, DBLP:conf/iccv/LinardosKPB21}.
In this paper, rather than utilizing these complicated multi-step or post-processing approaches, we undertake the calibration of the source model through the direct optimization of the ECE score. Additionally, we estimate and leverage ECE to enhance the source model's capacity for generalization to unseen domains.

\vspace{-2ex}
\section{Method}\vspace{-1ex}
\subsection{Notations and Definitions}
In domain adaptive semantic segmentation, the source domain $\mathcal{D}_S$ consists of $N_S$ labeled images $\mathcal{D}_S = \{x_i, y_i\}_{i=1}^{N_S}$, while the target set only includes the unlabeled images $\mathcal{D}_T = \{x_j\}_{j=1}^{N_T}$ of size $N_T$. Each image is represented as $x_i \in \mathbb{R}^{H \times W \times 3}$ and the label map as $y_i\in\{(0,1)\}^{H \times W \times C}$, where $H$, $W$ and $C$ represent the height, width, and number of classes, respectively. \vspace{-2ex}

\subsection{Overview}\vspace{-1ex}
In this work, we aim to answer two key questions in the source-free scenario: (1) What kind of source pre-trained model can benefit target adaptation? (2) How to ensure faithful pseudo labels during self-training on the target domain? To discover the answers, in Sec \ref{Differentiable-ECE}, we propose a method to generalize the source model $f(\cdot;\theta)$ by optimizing the expected calibration error (ECE) during the source pretraining phase. We further explain in Sec \ref{source-model-selection} that ECE is also an effective metric to ensure the generalizability in source model selection. To find out a proper threshold for pseudo-labeling, we introduce the value net $v(\cdot;\phi)$ in Sec \ref{value-net} to estimate the ECE of the unlabeled target data and incorporate this estimation into selecting the confidence threshold and self-training process in Sec \ref{target-adapt}. Finally, in Sec \ref{tar-sel}, a target model selection strategy is considered without access to the target label.
\vspace{-2ex}
\subsection{Differentiable Expected Calibration Error} 
During the source pretraining, our goal is to obtain a \textit{generalizable} source model with source data discriminative power to the target domain. To achieve this, as illustrated in Figure \ref{fig:src}, we train the source model in a supervised manner with the optimization of ECE. The objectives of model training on the source data are defined as:
\begin{equation}\label{eq:src}
    \mathcal{L}_{\operatorname{src}} = \mathcal{L}_{\text{seg}} + \alpha\mathcal{L}_{\operatorname{ECE}_\text{diff}},
\end{equation}
where $\alpha$ is a loss coefficient. The first term $\mathcal{L}_{\text{seg}}$ is to ensure the effectiveness of the model per pixel on the labeled source data. It is typically a pixel-level cross-entropy loss:
\begin{equation}\label{CE}
\mathcal{L}_{\text{seg}}=\sum_{(x, y) \in\mathcal{D}_S} -y \log f(x; \theta).
\end{equation}
After a long-time training, this could lead to overfitting of the training data (\textit{i.e., } the source data) \cite{DBLP:series/lncs/Prechelt12} and negatively impact the model's generalization ability on test data (\textit{i.e.,} the target data).

To avoid overfitting on the source domain, we propose to directly optimize \textbf{expected calibration error (ECE)}. ECE is a metric that quantifies the absolute difference between the accuracy and confidence of a model, where a lower ECE indicates better calibration. To calculate ECE, the confidence values of pixel-level predictions are split into $M$ bins of fixed intervals from 0 to 1. For each bin, the absolute difference between accuracy and the corresponding predicted confidence is computed as:
\begin{equation}\label{eq:L_ece}
\mathcal{L}_{\operatorname{ECE}} = \text{ECE}=\sum_{m=1}^M \frac{\left|B_m\right|}{n}\left|\operatorname{acc}\left(B_m\right)-\operatorname{conf}\left(B_m\right)\right|,
\end{equation}
where $n$ is the number of samples.  $\operatorname{acc}\left(B_m\right)$ and $\operatorname{conf}\left(B_m\right)$ are the accuracy and average confidence for the $m^{th}$ bin $B_m$, respectively.  

\subsubsection{\textbf{ECE Optimization}}\label{Differentiable-ECE} In particular, the confidence scores are obtained by finding the highest probabilistic outcome from each sample. For a sample $x$:
\begin{equation}
    \operatorname{conf}(x) = \operatorname{max}(f(x; \theta)),
\end{equation}
Note that the non-differentiable operation $\text{max}(\cdot)$ is applied in the calculation of average confidence, which poses a challenge when we expect to directly optimize $\mathcal{L}_{\operatorname{ECE}}$ in Eq. \eqref{eq:L_ece}. To address this issue, we propose a differentiable alternative $\mathcal{L}_{\operatorname{ECE}_\text{diff}}$ to enable ECE optimization during training on the source data. The main idea is to use a continuous simulation to approximate the max operation. To this end, we adopt the \textbf{LogSumExp (LSE)} function to smoothly close to the maximum function when calculating the confidence: 
\begin{equation}
\operatorname{LogSumExp}\left(\cdot \right)=t \cdot \log \sum_{i=1}^C e^{\frac{z_i}{t}},
\end{equation}
where $z_i$ is the $i^{th}$ element of the model output $f(x; \theta)$, $t$ is a scaling factor to control the ability of LSE to approximate the $max(\cdot)$ operation, which is set to $1e^{-5}$ empirically.

\subsubsection{\textbf{ECE Model Selection}} \label{source-model-selection}
Once the segmentation model is converged with respect to both the $\mathcal{L}_\text{seg}$ and $\mathcal{L}_{\operatorname{ECE}_\text{diff}}$ metrics, a generalizable source model can in principle benefit the target adaptation process. However, due to the non-coexistence of the source and target data, how to select the optimal checkpoint remains unclear. To this end, we propose a \textit{fair} model selection strategy for source checkpoints, motivated by the positive correlation between the model calibration performance and out-of-domain generalization capacity \cite{DBLP:conf/nips/MindererDRHZHTL21}. \eat{Initially, we randomly sample a small subset of the source data for model selection, denoted as $\mathcal{D}_{\operatorname{sub}} \subset \mathcal{D}_S$. }For each checkpoint with a set of parameters in the checkpoint pool, $\theta_i \in \Theta$, we evaluate the ECE scores for all randomly sampled validation images and select the best checkpoint $\theta^*$ based on the following criteria:
\begin{equation}
\theta^*=\underset{\theta_i \in \Theta}{\operatorname{argmin}} (\text{mean}(\text{ECE}_{\theta_i}) + \text{max}(\text{ECE}_{\theta_i}) + \text{min}(\text{ECE}_{\theta_i})),
\end{equation}
where we ensure no peak and valley can fluctuate too much of the model calibration. 

\subsubsection{\textbf{ECE Estimation}} \label{value-net}
Model calibration can be easily accomplished in supervised learning tasks with labeled data. However, in the context of SFDA, where the target labels are unavailable, implementing model calibration on the target side becomes challenging. To address this, our solution is to train a value net $v(\cdot;\phi)$ parameterized by $\phi$ to predict the ECE of the source data so that the value net can be reused to estimate ECE for the target data in the next stage. Specifically, we first freeze the segmentation network with selected parameter $\theta ^*$. Then, we treat the feature representation from the intermediate layer $l$ of the segmentation model as the input of the value net. The objective is to force each image's output of the value net to closely match the ground-truth ECE. Here, we  minimize the difference between the ground-truth ECE and the value net output $\widehat{\text{ECE}} = v(f_l(x;\theta^*); \phi)$ with Mean Squared Error (MSE) by:
\begin{equation}
     \mathcal{L}_{\text{match}}=\frac{1}{N_S} \sum_{i=1}^{N_S} \lVert \text{ECE}_i-\widehat{\text{ECE}}_i\rVert_2^2,
\end{equation}
where $N_S$ is the number of image samples in the source domain. The ground-truth ECE is calculated per image, and $\widehat{\text{ECE}}_i$ is the estimated ECE score for the $i$-th source image. 

\begin{figure*}[ht]
\centering\vspace{-2ex}
\includegraphics[width=1\linewidth]{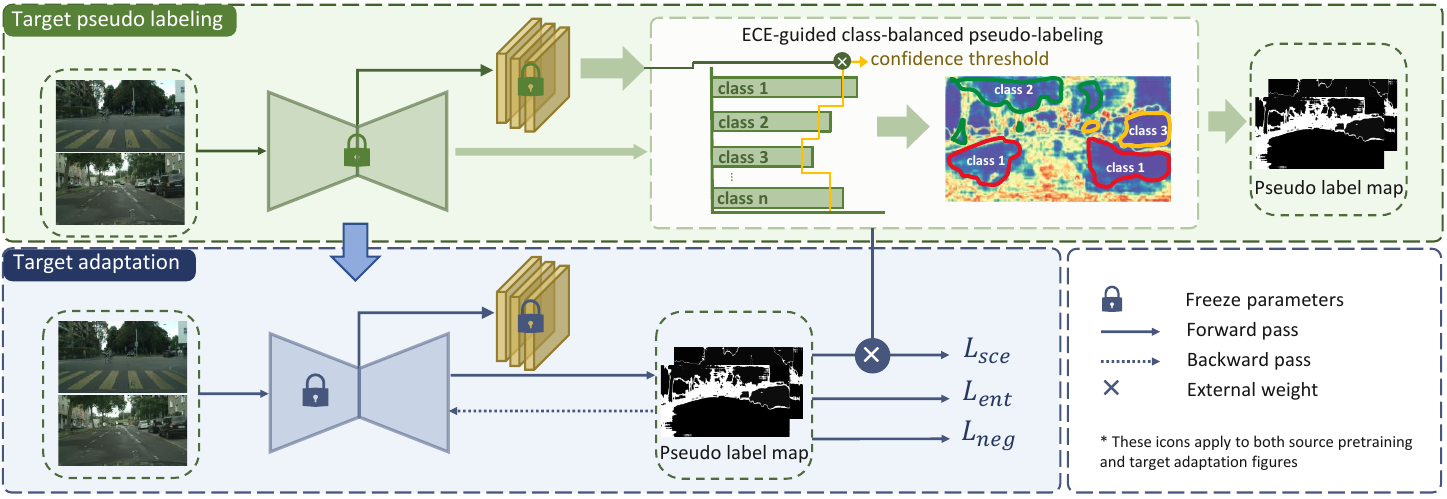}\vspace{-1ex}
\caption{An illustration of calibration-aware self-training during the target adaptation stage. Confident yet unreliable (\textit{i.e.,} of high estimated ECE scores) samples will be down-weighted, making it less likely to be pseudo-labeled and thus preventing potential error accumulation.\vspace{-2ex}}\label{fig:target}
\end{figure*}

\subsection{Calibration-aware Target Adaptation} \label{target-adapt}
Due to the unavailability of the target annotation, \textit{self-training} seems to be an indispensable method for adapting a model to the target domain in semantic segmentation tasks. Previous self-training approaches primarily pseudo-label the target data with \textit{high confidence} irrespective of its classes. However, in the case of semantic segmentation, this approach has resulted in tail classes being \textit{underrepresented} and biased pseudo-labels. For instance, if the threshold is set to $0.5$, a tail class with a maximum confidence score of $0.4$ will not be labeled. As a result, it will never be learned during the target adaptation process. Additionally, the confidence scores generated by the source model may not accurately reflect target performance, leading to erroneous pseudo-labels and accumulated incorrectness during adaptation. As shown in Figure \ref{fig:target}, our proposed calibration-aware self-training method explicitly addresses the limitations of the conventional approach with the following three steps:

\subsubsection{\textbf{Reliable Class-based Pseudo-labeling}}
 To mitigate the issues posed in original pseudo-labeling, we present a novel method called reliable class-based pseudo-labeling, which labels the same proportion of data across all classes as \cite{DBLP:conf/eccv/LiKLWY20} under the guidance of the estimated Expected Calibration Error $\widehat{\text{ECE}}$ from the value net $v$. 

To be specific, for pixels predicted to be class $c$, their confidence is denoted as $p^c$. However, $p^c$ may be significantly biased due to domain shift and imbalanced data \cite{DBLP:conf/cvpr/LiLHZJZ22}. Therefore, we utilize a calibration-guided confidence score $P^c$ to improve the thresholding and pseudo-labeling process:
\begin{equation}
    P^c =(1-\widehat{\text{ECE}}) \cdot p^c.
\end{equation}
The intuition behind it is that a lower ECE indicates better consistency between accuracy and confidence scores. Therefore, for the target images with low $\widehat{\text{ECE}}$, we assume that these images are well-calibrated and require little modification to the pixel-level confidence score. Conversely, if an image has a high $\widehat{\text{ECE}}$, it may have low confidence reliability, so we scale down its confidence value to avoid potential overfitting and reduce the likelihood of being selected as a pseudo-label. Furthermore, we group the calibration-guided confidence score $P^c$ into class confidence set $\mathfrak{R}^c$ for each class $c$ in descending order:
\begin{equation}
    \mathfrak{R}^c= \{ P^c_1, P^c_2, \cdots , P^c_m\},
\end{equation}
where $m$ is the total number of samples to be predicted as class $c$. For each element in $\mathfrak{R}^c$, we have $P^c_i \geq P^c_{i+1}$. The smallest calibration-guided confidence score in $\mathfrak{R}^c$ is $ P^c_m$. A hyperparameter $\delta \in [0,1]$ is then introduced as the ratio for selecting pseudo-labels. The confidence threshold for class $c$ is calculated by:
\begin{equation}
    \xi _c = \mathfrak{R}^c[\delta \cdot | \mathfrak{R}^c |],
\end{equation}
which means the top $\delta$ of the data belong to class $c$ will be pseudo-labeled. $\delta \cdot | \mathfrak{R}^c |$ is the index of the one to be the threshold. We apply the same strategy as \cite{DBLP:conf/eccv/LiKLWY20} at the image level and empirically set $\delta$ to $15\%$ for all experiments. For a pixel $x_i \in \mathcal{D}_T$, if it is predicted as class $c$, its pseudo-label assignment is defined as:
\begin{equation}
    \hat{y}_i =\begin{cases}
		c, & \text{if }~ f(x_i; \theta) > \xi _c\\
            255.   & \text{otherwise},
		 \end{cases}
\end{equation}
where $255$ refers to the background class that will not join the adaptation process.

By incorporating $\widehat{\text{ECE}}$ in the threshold calculation, our method enables a balance between the accuracy of the pseudo-labels and the confidence reliability. This potentially improves the calibration of the pseudo-labels and facilitates the adaptation of the source model to the target domain, as discussed in Sec \ref{Ablation}. Additionally, using the same selection ratio $\delta$ for all classes alleviates the issue of optimizing tail classes that lack class pseudo-labels.

\noindent\textbf{Statistic Warm-up.}

To ensure stable adaptation during self-training, instead of using source statistics directly, we freeze the whole model except the BatchNorm for the first epoch of each new round and update the normalization statistics mean $\mu$ and the variance $\sigma^2$ of the activation values for the current batch:
\begin{equation}
    \mu \leftarrow \mathbb{E}\left[x_b\right],   \sigma^2 \leftarrow \mathbb{E}\left[\left(\mu-x_b\right)^2\right],
\end{equation}
where $x_b$ is the batch $b$ of data pixels.
More details can be found in the \textit{appendix}.

\subsubsection{\textbf{Weighted Self-training}}\label{w-sce} 
After the statistic warm-up, we optimize a symmetric cross-entropy \cite{DBLP:conf/iccv/0001MCLY019} weighted by normalized class-level thresholds:
\begin{equation}
    \mathcal{L}_{\text {sce }} = \epsilon\mathcal{L}_{\text {wCE }} + \mathcal{L}_{\text {rCE }},
\end{equation}
where  $\epsilon$ is a loss coefficient. $\mathcal{L}_{\text {wCE}}$ is the weighted cross-entropy loss conducted on target data and their corresponding pseudo-labels, $\mathcal{L}_{\text{rCE}}$ is the reversed version of cross-entropy loss, with exchanged prediction and pseudo-labels. They can be formulated as:
\begin{equation}
    \mathcal{L}_{\text {wCE }}=  - \sum_{(x, \hat{y}) \in\left(\mathcal{X}_T^{\prime}, \hat{\mathcal{Y}}_T^{\prime}\right)}  w \hat{y} \log f(x; \theta),
\end{equation}
\begin{equation}
    \mathcal{L}_{\text {rCE }} = - \sum_{(x, \hat{y}) \in\left(\mathcal{X}_T^{\prime}, \hat{\mathcal{Y}}_T^{\prime}\right)}  f(x; \theta) \log \hat{y}.
\end{equation}
where $w$ is a $C$-dimensional weight metric with $ w_c = \frac{\exp \left(\xi_c\right)}{\sum_{j=1}^C \exp \left(\xi_j\right)}$ in dimension $c$.  $\mathcal{X}_T^{\prime}$ is the set of pseudo-labeled data without background (\textit{i.e.,} pixels being labeled as 255), $\hat{\mathcal{Y}}_T^{\prime}$ is the corresponding pseudo-label set. By applying $\mathcal{L}_\text{sce}$, the data noises that come from pseudo-labeling could be significantly mitigated.

An entropy loss is then employed as a regularizer:
\begin{equation}
\mathcal{L}_\text{ent} = -\sum_{x\in \mathcal{X}_T} f(x; \theta) \log(f(x; \theta))
\end{equation}

To gain more information about which class the data belongs to, negative learning \cite{DBLP:conf/iccv/KimYYK19} is considered a typical way of dealing with SFDA for segmentation:
\begin{equation}
    \mathcal{L}_{\text {neg }}= - \sum_{(x, \bar{y}) \in\left(\mathcal{X}_T^{\prime}, \bar{\mathcal{Y}}_T^{\prime}\right)} \bar{y} \log(1- f(x; \theta)),
\end{equation}
where $\bar{y}$ is the class that not equal to the pseudo-label $\hat{y}$.

To conclude, the overall objective of the self-training process is:
\begin{equation}
    \mathcal{L}_{\text {tar }} = \mathcal{L}_{\text {sce }}  + \mathcal{L}_{\text {neg}} + \eta \mathcal{L}_{\text {ent }}
\label{eq:ltar}
\end{equation}
where $\eta$ is the loss coefficient.
\subsubsection{\textbf{Discussion on target model selection}}\label{tar-sel}
After fine-tuning the source model on the target data, a new challenge arises: \textit{How to select the best model without the presence of target labels?} The conventional method entails partitioning the target dataset into training and testing subsets and utilizing the testing performance as the basis for model selection, which violates the assumption that the target domain does not contain any labels. This paper adopts a model selection method that does not require any labels from the target domain.
Based on the findings in \cite{DBLP:journals/corr/abs-2302-13824}, the sum of model uncertainty corresponds to the entropy metric; thus, a lower entropy implies a higher degree of certainty for the model. More details and discussions can be found in Sec \ref{abl:target-model-selection}.

\begin{table*}[t]
\centering \vspace{-3ex}
\caption{Results of GTA5$\rightarrow$Cityscapes.\vspace{-3ex}}
\resizebox{1\linewidth}{!}{%
\begin{tabular}{l l c c c c c c c c c c c c c c c c c c c | c}
\toprule
\multirow{1}{*}{SF} &\multirow{1}{*}{Method} &{\rotatebox{90}{road}} &{\rotatebox{90}{side.}} &{\rotatebox{90}{build.}} &{\rotatebox{90}{wall$^*$}} &{\rotatebox{90}{fence$^*$}} &{\rotatebox{90}{pole$^*$}} &{\rotatebox{90}{light}} &{\rotatebox{90}{sign}} &{\rotatebox{90}{vege.}} &{\rotatebox{90}{terr.}}  &{\rotatebox{90}{sky}} &{\rotatebox{90}{pers.}} &{\rotatebox{90}{rider}} &{\rotatebox{90}{car}} &{\rotatebox{90}{truck}} &{\rotatebox{90}{bus}} &{\rotatebox{90}{train}} &{\rotatebox{90}{motor}} &{\rotatebox{90}{bike}} &{\rotatebox{90}{mIoU}}\\
\midrule 
\midrule
\multirow{8}{*}{{\ding{55}}} 
 &SIBAN \cite{luo2019significance} &88.5 &35.4 &79.5 &26.3 &24.3 &28.5 &32.5 &18.3 &81.2 &40.0 &76.5 &58.1 &25.8 &82.6 &30.3 &34.4 &3.4 &21.6 &21.5 &42.6\\
 &AdaptSeg \cite{DBLP:conf/cvpr/TsaiHSS0C18} &87.3 &29.8 &78.6 &21.1 &18.2 &22.5 &21.5 &11.0 &79.7 &29.6 &71.3 &46.8 &6.5 &80.1 &23.0 &26.9 &0.0 &10.6 &0.3 &35.0\\
 &CLAN \cite{luo2021category} &88.7 &35.5 &80.3 &27.5 &25.0 &29.3 &36.4 &28.1 &84.5 &37.0 &76.6 &58.4 &29.7 &81.2 &38.8 &40.9 &5.6 &32.9 &28.8 &45.5\\
 &DPR \cite{DBLP:conf/iccv/TsaiSSC19} &92.3 &51.9 &82.1 &29.2 &25.1 &24.5 &33.8 &33.0 &82.4 &32.8 &82.2 &58.6 &27.2 &84.3 &33.4 &46.3 &2.2 &29.5 &32.3 &46.5\\
 &IntraDA \cite{DBLP:conf/cvpr/PanSRLK20} &90.6 &37.1 &82.6 &30.1 &19.1 &29.5 &32.4 &20.6 &85.7 &40.5 &79.7 &58.7 &31.1 &86.3 &31.5 &48.3 &0.0 &30.2 &35.8 &46.3\\
 &CRST \cite{DBLP:conf/iccv/ZouYLKW19} &91.0 &55.4 &80.0 &33.7 &21.4 &37.3 &32.9 &24.5 &85.0 &34.1 &80.8 &57.7 &24.6 &84.1 &27.8 &30.1 &26.9 &26.0 &42.3 &47.1\\
 &DAST \cite{DBLP:conf/aaai/YuZDHDZ21} &92.2 &49.0 &84.3 &36.5 &28.9 &33.9 &38.8 &28.4 &84.9 &41.6 &83.2 &60.0 &28.7 &87.2 &45.0 &45.3 &7.4 &33.8 &32.8 &49.6\\
 &CCM \cite{DBLP:conf/eccv/LiKLWY20} &93.5 &57.6 &84.6 &39.3 &24.1 &25.2 &35.0 &17.3 &85.0 &40.6 &86.5 &58.7 &28.7 &85.8 &49.0 &56.4 &5.4 &31.9 &43.2 &49.9\\
 
\midrule
\multirow{9}{*}{{\ding{51}}} 
&Source only ECE &65.9 &13.5 &64.9 &20.7 &19.8 &28.5 &35.4 &30.8 &81.1 &10.2 &65.3 &59.2 &23.9 &75.9 &31.0 &18.5 &2.0 &22.7 &41.7 & 37.4\\ 
&EntMin \cite{DBLP:conf/cvpr/VuJBCP19} &82.8 &0.0 &70.2 &2.2 &0.3 &0.4 &2.8 &1.6 &79.9 &8.1 &79.2 &22.2 &0.1 &83.1 &22.5 &30.0 &2.0 &6.3 &0.0 &26.0\\
&Pseudo &83.2 &0.0 &67.3 &1.1 &0.0 &0.1 &1.2 &1.2 &77.7 &1.3 &81.4 &11.5 &0.0 &81.7 &18.0 &14.8 &0.0 &3.7 &0.0 &23.4\\
&Pse.+Ent. &83.0 &0.0 &66.0 &0.2 &0.0 &0.0 &0.4 &0.5 &75.4 &0.0 &82.3 &7.3 &0.0 &80.8 &12.5 &2.6 &0.0 &2.2 &0.0 &21.8\\
&Pse.+Sel. &83.2 &0.8 &76.1 &13.5 &7.9 &4.4 &9.2 &5.9 &82.9 &27.3 &77.2 &41.0 &1.8 &83.8 &36.3 &45.8 &5.0 &15.8 &0.0 &32.5\\
&SHOT \cite{DBLP:conf/icml/LiangHF20} &87.6 &44.4 &80.6 &24.4 &19.4 &9.8 &14.4 &9.6 &83.5 &37.6 &79.8 &49.6 &0.0 &78.6 &36.7 &50.1 &8.0 &18.0 &0.0 &38.5\\
&S4T \cite{DBLP:journals/corr/abs-2107-10140} &89.7 &46.7 &84.4 &25.7 &29.0 &39.5 &45.1 &36.8 &86.8 &41.8 &79.3 &61.2 &26.7 &85.0 &19.3 &28.2 &5.3 &11.8 &9.3 &44.8\\
&LD \cite{DBLP:conf/mm/YouLZCH21} &91.6 &53.2 &80.6 &36.6 &14.2 &26.4 &31.6 &22.7 &83.1 &42.1 &79.3 &57.3 &26.6 &82.1 &41.0 &50.1 &0.3 &25.9 &19.5 &45.5\\
&DTAC \cite{DBLP:conf/icmcs/YangKH22} &78.0 &29.5 &83.0 &29.3 &21.0 &31.8 &38.1 &33.1 &83.8 &39.2 &80.8 &61.0 &30.0 &83.9 &26.1 &40.4 &1.9 &34.2 &43.7 &45.7\\

\midrule
\midrule

 &Cal-SFDA &90.0 &\textbf{48.4} &83.2 &35.5 &23.6 &30.8 &39.6 &35.9 &84.3 &\textbf{43.2} &\textbf{85.1} &60.2 &27.9 &84.3 &32.6 &44.7 &2.2 &19.9 &42.1 &\textbf{48.1}\\

\bottomrule
\end{tabular}}\vspace{-2ex}
\label{gta5}
\end{table*}

\begin{table*}[t]
\centering 
\caption{Results of Synthia$\rightarrow$Cityscapes. mIoU$^*$ is the mean IoU of 13 classes, excluding the classes marked with `$^*$'. \vspace{-3ex}}
\resizebox{1\linewidth}{!}{%
\begin{tabular}{l l c c c c c c c c c c c c c c c c | c c}

\toprule
\multirow{1}{*}{SF} &\multirow{1}{*}{Method} &{\rotatebox{90}{road}} &{\rotatebox{90}{side.}} &{\rotatebox{90}{build.}} &{\rotatebox{90}{wall$^*$}} &{\rotatebox{90}{fence$^*$}} &{\rotatebox{90}{pole$^*$}} &{\rotatebox{90}{light}} &{\rotatebox{90}{sign}} &{\rotatebox{90}{vege.}} &{\rotatebox{90}{sky}} &{\rotatebox{90}{pers.}} &{\rotatebox{90}{rider}} &{\rotatebox{90}{car}} &{\rotatebox{90}{bus}} &{\rotatebox{90}{motor}} &{\rotatebox{90}{bike}} &{\rotatebox{90}{mIoU}} &{\rotatebox{90}{mIoU$^*$}}\\
\midrule 
\midrule
\multirow{8}{*}{{\ding{55}}} 
 &SIBAN \cite{luo2019significance} &82.5 &24.0 &79.4 &- &- &- &16.5 &12.7 &79.2 &82.8 &58.3 &18.0 &79.3 &25.3 &17.6 &25.9 &- &46.3\\
 &AdaptSeg \cite{DBLP:conf/cvpr/TsaiHSS0C18} &84.3 &42.7 &77.5 &- &- &- &4.7 &7.0 &77.9 &82.5 &54.3 &21.0 &72.3 &32.2 &18.9 &32.3 &- &46.7\\
 &CLAN \cite{luo2021category} &82.7 &37.2 &81.5 &- &- &- &17.1 &13.1 &81.2 &83.3 &55.5 &22.1 &76.6 &30.1 &23.5 &30.7 &- &48.8\\
 &DPR \cite{DBLP:conf/iccv/TsaiSSC19} &82.4 &38.0 &78.6 &8.7 &0.6 &26.0 &3.9 &11.1 &75.5 &84.6 &53.5 &21.6 &71.4 &32.6 &19.3 &31.7 &40.0 &46.5\\
 &IntraDA \cite{DBLP:conf/cvpr/PanSRLK20} &84.3 &37.7 &79.5 &5.3 &0.4 &24.9 &9.2 &8.4 &80.0 &84.1 &57.2 &23.0 &78.0 &38.1 &20.3 &36.5 &41.7 &48.9\\
 &CRST \cite{DBLP:conf/iccv/ZouYLKW19} &67.7 &32.2 &73.9 &10.7 &1.6 &37.4 &22.2 &31.2 &80.8 &80.5 &60.8 &29.1 &82.8 &25.0 &19.4 &45.3 &43.8 &50.1\\
 &DAST \cite{DBLP:conf/aaai/YuZDHDZ21} &87.1 &44.5 &82.3 &10.7 &0.8 &29.9 &13.9 &13.1 &81.6 &86.0 &60.3 &25.1 &83.1 &40.1 &24.4 &40.5 &45.2 &52.5\\
 &CCM \cite{DBLP:conf/eccv/LiKLWY20} &79.6 &36.4 &80.6 &13.3 &0.3 &25.5 &22.4 &14.9 &81.8 &77.4 &56.8 &25.9 &80.7 &45.3 &29.9 &52.0 &45.2 &52.9\\
 
\midrule
\multirow{9}{*}{{\ding{51}}} 

&Source only ECE &53.4 &21.6 &77.3 &7.4 &0.4 &26.2 &20.0 &14.6 &75.9 &78.3 &50.0 &20.6 &59.4 &28.6 &23.6 &38.1  &37.2 &43.2 \\ 
 
&EntMin \cite{DBLP:conf/cvpr/VuJBCP19} &80.6 &0.3 &72.4 &0.4 &0.0 &3.7 &0.4 &3.4 &73.2 &72.3 &20.6 &4.2 &78.6 &23.0 &1.2 &0.0 &27.1 &33.1\\
&Pseudo &82.6 &0.0 &66.5 &0.0 &0.0 &0.3 &0.0 &0.4 &69.0 &74.5 &0.4 &0.5 &79.8 &0.4 &0.4 &0.0 &23.4 &28.8\\
&Pse.+Ent. &81.8 &0.0 &68.5 &0.0 &0.0 &0.3 &0.0 &0.6 &72.0 &75.1 &1.2 &0.6 &79.4 &0.4 &0.6 &0.0 &23.8 &29.2\\
&Pse.+Sel. &80.6 &2.4 &75.5 &2.2 &0.0 &11.5 &1.1 &7.7 &75.5 &72.9 &40.0 &9.7 &80.0 &44.1 &3.9 &1.1 &31.8 &38.0\\
&SHOT \cite{DBLP:conf/icml/LiangHF20} &61.3 &26.4 &74.7 &5.1 &0.0 &18.8 &0.0 &20.9 &75.6 &63.6 &14.5 &0.0 &52.0 &34.0 &2.2 &1.6 &28.2 &32.8\\
&S4T \cite{DBLP:journals/corr/abs-2107-10140} &84.9 &43.2 &79.5 &7.2 &0.3 &26.3 &7.8 &11.7 &80.7 &82.4 &52.4 &18.7 &77.4 &9.6 &9.5 &37.9 &39.3 &45.8\\
&LD \cite{DBLP:conf/mm/YouLZCH21} &77.1 &33.4 &79.4 &5.8 &0.5 &23.7 &5.2 &13.0 &81.8 &78.3 &56.1 &21.6 &80.3 &49.6 &28.0 &48.1 &42.6 &50.1\\
&DTAC \cite{DBLP:conf/icmcs/YangKH22} &77.5 &37.4 &80.5 &13.5 &1.7 &30.5 &24.8 &19.7 &79.1 &83.0 &49.1 &20.8 &76.2 &12.1 &16.5 &46.1 &41.8 &47.9\\

\midrule
\midrule
 &Cal-SFDA
 &76.3 &32.6 &\textbf{81.2} &4.0 &0.6 &27.5 &20.2 &17.6 &\textbf{82.4} &\textbf{83.1} &51.8 &18.1 &\textbf{83.3} &46.2 &14.7 &\textbf{48.1} &\textbf{43.0} &\textbf{50.4}\\
\bottomrule
\end{tabular}}\vspace{-2ex}
\label{syn}
\end{table*}
\vspace{-2ex}
\section{Experiments}

\vspace{-1ex}
\subsection{Experimental details} \label{exp}
\textbf{Datasets. }We assess the efficacy of our proposed approach in synthetic-to-real scenarios, which are known to be challenging. Specifically, we train our models on fully-annotated synthetic data and then adapt and evaluate them on real-world data. We use two commonly used synthetic datasets, namely GTA5 \cite{DBLP:conf/eccv/RichterVRK16} and SYNTHIA \cite{DBLP:conf/cvpr/RosSMVL16}, along with the Cityscapes \cite{DBLP:conf/cvpr/CordtsORREBFRS16} dataset as the target domain data to evaluate the effectiveness of our method. 

\noindent\textbf{Network architectures.} We build our method upon the Deeplab-V2 \cite{DBLP:journals/pami/ChenPKMY18} architecture with a pre-trained ResNet-101 \cite{DBLP:conf/cvpr/HeZRS16}. The output of the last layer is subjected to the Atrous Spatial Pyramid Pooling (ASPP) module with preset sampling rates of 6, 12, 18, and 24. 

\noindent\textbf{Implementation details. } We use an up-sampling layer with a softmax operator to generate a segmentation map of the same size as the original image. The models are implemented using Pytorch and trained on a single NVIDIA RTX A6000 GPU. The segmentation network is trained using the Stochastic Gradient Descent (SGD) optimizer, with a momentum of 0.9 and weight decay of $5 \times 10^{-4}$. A polynomial learning rate scheduling is employed with a power of 0.9, an initial learning rate of $5 \times 10^{-4}$, and a batch size of 4. During source pretraining, we set the loss coefficient $\alpha$ as $1$. Following the hyperparameter set in previous works, $\epsilon$ is set to $0.1$, and $\eta$ is $0.005$ during target adaptation. See more details in the \textit{appendix}.

\vspace{-2ex}
\subsection{Experimental Results} \label{result}
We evaluate the efficacy of our proposed model using two widely used synthetic-to-real benchmarks, namely GTA5→Cityscapes, and Synthia→Cityscapes. To ensure a fair comparison, we conduct all experiments using the Deeplab v2 \cite{DBLP:journals/pami/ChenPKMY18} and pre-trained Resnet101 \cite{DBLP:conf/cvpr/HeZRS16}. The baselines that we primarily compare include S4T \cite{DBLP:journals/corr/abs-2107-10140}, LD \cite{DBLP:conf/mm/YouLZCH21}, and DTAC \cite{DBLP:conf/icmcs/YangKH22}, which all employ the ``source pretraining $+$ target adaptation'' process. All other experimental results are obtained from \cite{DBLP:conf/mm/YouLZCH21}.
\\
\textbf{GTA5$\rightarrow$Cityscapes: }
Table \ref{gta5} shows the experimental result of the GTA5$\rightarrow$Cityscapes benchmark, where the following observations can be made:
1) The source model with $\mathcal{L}_{\operatorname{ECE}_\text{diff}}$ demonstrates significantly better results compared to self-training based purely on confidence, which implies that source calibration not only enhances the reliability of the model (\textit{i.e.,} confidence aligns with accuracy) as shown in Table \ref{table:ablation} but also improves its generalizability.
2) After target adaptation, Cal-SFDA outperforms the previous state-of-the-art method DTAC by $5.25\%$ in terms of mIoU. Moreover, compared to LD which also employs negative learning, Cal-SFDA achieves a $5.71\%$ improvement. The proposed approach could also surpass S4T, which borrows the idea of augmentation consistency for more than $7.36\%$ mIoU. These results demonstrate that the calibration-guided learning approach we adopted for the target enhances the performance of the adaptation. However, it is worth noting that not every class has shown improved performance -  for instance, the ``trains'' class does not perform well, which could be attributed to the inconsistency in the class definition and annotation strategy between the source and target.\\
\textbf{Synthia$\rightarrow$Cityscapes: }
Table \ref{syn} presents the performance results for the 13- and 16-class cases. Our observations are consistent with those of GTA5, which are listed as follows:
1) Our proposed approach achieves state-of-the-art performance in terms of mean Intersection over Union (mIoU) for both the 13- and 16-class scenarios.
2) Through a comparison of the impact of various source domains on the final transfer outcomes, we note that, despite being virtual world datasets, GTA5 and Synthia might not demonstrate the same degree of domain shift as the real world, owing to factors such as the virtual environment's color and design style. Additionally, our ablation study enables us to understand better the influence of the different components proposed in this study on the overall performance. See Table \ref{table:ablation} for more results.\vspace{-2ex}

\begin{figure}[t]
    \centering\vspace{-2em}
    \setlength{\abovecaptionskip}{0.cm}
    \subfloat[][Model performance]{\includegraphics[width=0.48\linewidth, height=3.2cm]{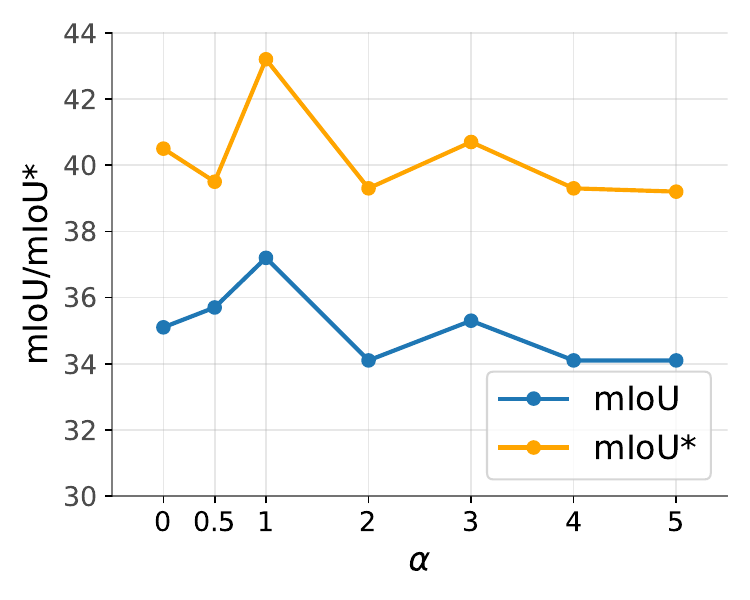}}
    \subfloat[][Model calibration]{\includegraphics[width=0.48\linewidth,height=3.2cm]{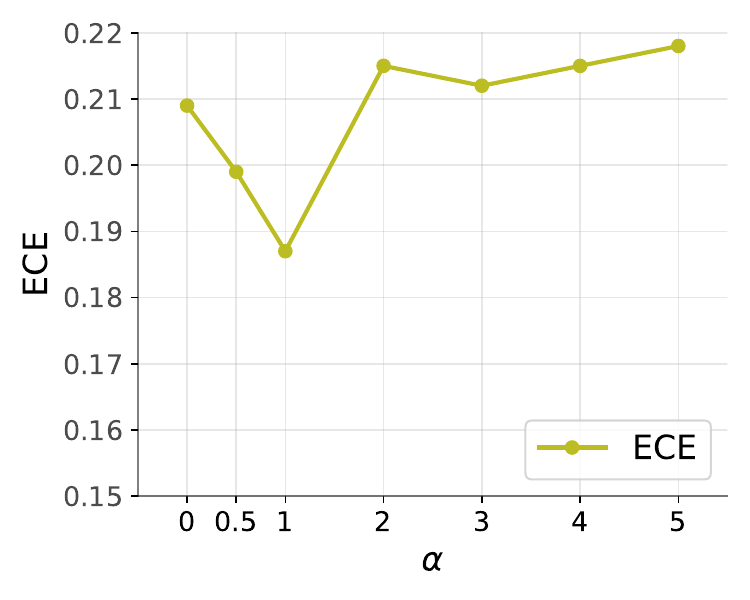}}
    \caption{Parameter sensitivity of loss coefficients $\alpha$ on during source model training.\vspace{-1ex}}
    \label{Figure:parameter-study}
    \vspace{-4ex}
\end{figure}
\vspace{-1ex}
\subsection{Parameter Sensitivity} \label{dis}
The present study aims to investigate the impact of hyperparameter $\alpha$ on a source model, specifically regarding the loss function $\mathcal{L}_{\operatorname{ECE}_\text{diff}}$. Our analysis considered two main aspects of varying $\alpha$: performance and calibration. $\alpha$ is defined as 0.5 and within a range from 0 to 5 with an interval of 1. The results of our experiments are presented in Figure \ref{Figure:parameter-study}. We first compare the calibration and non-calibration cases by setting $\alpha$ from 0 to 1. Our findings indicate that optimizing ECE in this range can enhance the model's performance on out-of-distribution data as well as the reliability of the model's confidence in the target domain, as evidenced by higher mIoU and lower ECE compared to the non-optimized model. However, increasing $\alpha$ results in a decline in model performance and increased ECE, which is attributed to the dominance of the original classification loss by the high weight of $\mathcal{L}_{\operatorname{ECE}_\text{diff}}$. Ultimately, we find that a weight of 1 yields the best response regarding mIoU and model calibration, possibly due to a trade-off between optimizing classification loss and maintaining model calibration. It is worth noting that all empirical results here are based on our proposed model selection strategy in Sec \ref{source-model-selection}. 

\begin{figure*}[ht]
\centering\vspace{-1ex}
\includegraphics[width=1\linewidth]{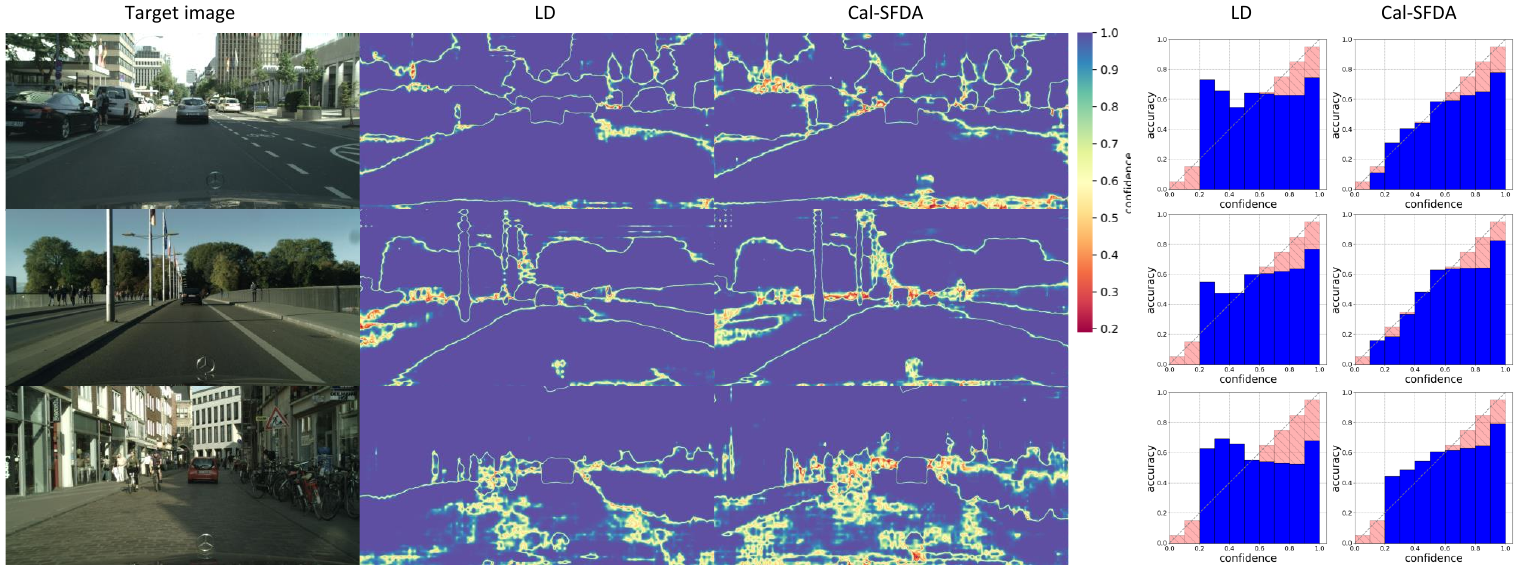}\vspace{-3ex}
\caption{Confidence map (Left) and the corresponding reliability diagram (Right) from LD\cite{DBLP:conf/mm/YouLZCH21} and our proposed Cal-SFDA.\vspace{-1ex}}\label{fig:cal-map}
\end{figure*}

\begin{figure*}[!htb]
\centering
\includegraphics[width=1\linewidth]{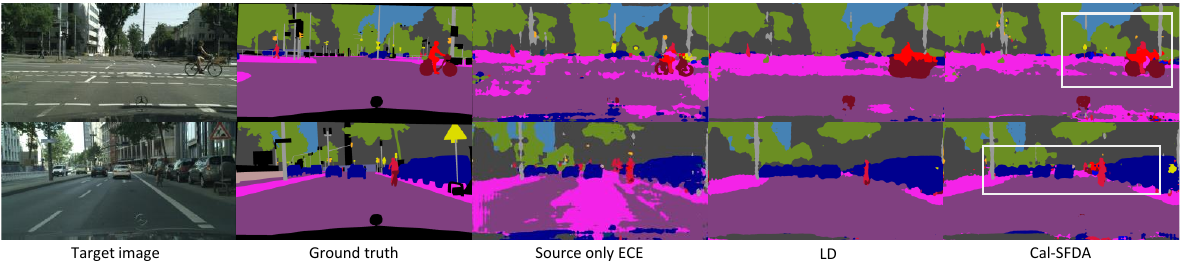}\vspace{-2ex}
\caption{Qualitative results of Cal-SFDA on Synthia→Cityscapes. White boxes are key areas.\vspace{-1ex}}\label{fig:label-map}
\end{figure*}

\subsection{Ablation Study} \label{ablation}

\subsubsection{\textbf{Impact of Component.}}
For an in-depth understanding of the contributions of individual components to the overall improvement, we conduct an ablation study. This involved systematically removing each core component, one at a time, and evaluating the resulting performance on the target test set.

\begin{table}[t]
\centering\vspace{-1em}
\caption{Ablation Study.\vspace{-3ex}}
\resizebox{0.96\linewidth}{!}{%
\begin{tabular}{c | c c c c c | c c c}
\midrule
Stage &ECE  & stat. & thres & w-ce & w-sce & mIoU $\uparrow$ & mIoU$^*$ $\uparrow$ &ECE $\downarrow$\\
\midrule
\midrule
\multirow{2}{*}{{S}}
 & & &  & & &35.1 &40.5 &0.209\\
 &\ding{51} &  &  & & &\textbf{37.2} &\textbf{43.2} &\textbf{0.187}\\
\midrule
\multirow{4}{*}{{T}}
 &\ding{51} & & &&  &40.0 &46.7 &0.178\\
 &\ding{51} & \ding{51} & & & &40.2 &47.0 &0.176\\
 &\ding{51} & \ding{51} & \ding{51}  & & &40.7 &48.0 &\textbf{0.161}\\
  &\ding{51} & \ding{51} & \ding{51} & \ding{51} & &40.5 &47.4 &0.166\\
 &\ding{51} & \ding{51} & \ding{51} & \ding{51} & \ding{51} &\textbf{43.0} &\textbf{50.4} &0.223\\
\midrule
\end{tabular}\vspace{-3em}}\vspace{-2ex}
\label{table:ablation}\vspace{-1ex}
\end{table}

Table \ref{table:ablation} summarizes the impact of each proposed component on the model's performance and calibration in terms of both the source training and target adaptation stages. Specifically, removing $\mathcal{L}_{\operatorname{ECE}_\text{diff}}$ from the source model (lines 1 and 2) results in a significant drop in performance, with mIoU decreasing by about $6\%$. The model's calibration is also affected, indicating that incorporating model calibration into the source model effectively improves its performance and calibration on out-of-distribution data.

 During target adaptation, our experiments showed that each of the proposed components benefits the final performance in terms of mIoU. Although the weighted cross-entropy degrades performance compared to the pure cross-entropy loss, it suggests that it may be too strong to be optimized with the weight. Furthermore, when weighted SCE is applied on top of these components, the model's performance improves from $40.5\%$ to $43.0\%$. However, the performance and calibration in the target domain are observed to be a trade-off when the target label is unavailable.
\vspace{-1ex}
\begin{table}[h]\vspace{-2ex}
\centering
\caption{Comparisons on source model selection.\vspace{-2ex}}
\resizebox{0.85\linewidth}{!}{%
\begin{tabular}{l | c c c}
\toprule
Strategy  & mIoU $\uparrow$ & mIoU$^*$ $\uparrow$ & ECE $\downarrow$ \\
\midrule
\midrule
Source Risk & 35.0 &40.2 &0.271 \\
 Ours & 37.2 &43.2 & 0.230 \\
 \midrule
 Target Risk (Upper Bound)  & 37.5 &42.6 &0.202 \\ 
\bottomrule
\end{tabular}}\vspace{-2.5ex}
\label{table:model-selection}
\end{table}
\subsubsection{\textbf{Impact of Source Model Selection.}} \label{abl:source-model-selection} In regards to selecting the source model parameters from the checkpoint pool $\Theta$, we evaluate three strategies: 1) selecting parameters based solely on the performance of the source data, 2) directly selecting the checkpoint with the highest target performance, which is not reasonable due to the unavailability of target data and labels during source pre-training; and 3) adopting our proposed strategy of selecting the checkpoint with the minimum ECE in sec \ref{source-model-selection} on the source data. The experiments presented in Table \ref{table:model-selection} show that the strategy relying exclusively on the target performance is the optimal choice. Considering the limitations imposed by the source-free and UDA settings, we compare the source performance and the source's lowest ECE and find that our proposed selection strategy achieves a higher target mIoU and a lower target ECE (\textit{i.e.,} better model calibration), which demonstrates the effectiveness of our strategy.
\begin{table}[h]
\centering\vspace{-2ex}
\caption{Target model selection.\vspace{-3ex}}
\label{Ablation}
\resizebox{0.85\linewidth}{!}{%
\vspace{-3ex}
\begin{tabular}{l | c c c}
\toprule
Strategy  & mIoU $\uparrow$ & mIoU$^*$ $\uparrow$ & ECE $\downarrow$ \\
\midrule
\midrule
 Nuclear Norm \cite{DBLP:journals/corr/abs-2107-06154} &42.5 &50.2 &0.231\\
 Ours &\textbf{43.0} &\textbf{50.4} &\textbf{0.223} \\
 \midrule
 Target Risk (Upper Bound) &\textbf{43.0} &\textbf{50.4} &\textbf{0.223} \\
\bottomrule
\end{tabular}}\vspace{-3ex}
\label{table:tar-model-selection}
\end{table}
\subsubsection{\textbf{Impact of Target Model Selection.}} \label{abl:target-model-selection}
This study uses Shannon entropy to identify the ultimate target checkpoint. We also compare its efficacy with other selection methods, including 1) choosing the checkpoint that has the greatest nuclear norm \cite{DBLP:journals/corr/abs-2107-06154}, as computed on the feature at $f_l$; and 2) directly selecting the parameters that result in the best target performance. The nuclear norm, calculated by the summation of the singular values of the feature map, is treated as a representative of choosing the target checkpoint. We expect a target checkpoint with the largest nuclear norm to ensure diversified target predictions. As table \ref{table:tar-model-selection} shows, the feature-level nuclear norm seems suboptimal. Nevertheless, minimizing the Shannon entropy yielded the best target performance, thereby validating the efficacy of our target model selection methodology.
\vspace{-2ex}
\subsection{Qualitative Study}
\textbf{Calibration Analysis. }
Figure \ref{fig:cal-map} displays the confidence map and the corresponding reliability diagram for three randomly selected images. Our goal is to ensure that the predicted confidence accurately reflects the uncertainty in each area rather than merely predicting each pixel with high confidence. Notably, Cal-SFDA has a soft margin for uncertain categories, as shown on the left side of Figure \ref{fig:cal-map}. Conversely, both the source-only ECE baseline (\textit{i.e., } the source pre-trained model with ECE optimization) and LD \cite{DBLP:conf/mm/YouLZCH21} lack a soft margin. Concerning confidence reliability, our methods achieve competitive results with fewer under- and overconfidence issues as the results are closer to the diagonal line depicted on the right-side reliability diagram of Figure \ref{fig:cal-map}. 
\\
\textbf{Segmentation Visualization. }
Figure \ref{fig:label-map} visualizes the predicted label map for our source-only ECE model, baseline LD, and proposed Cal-SFDA. Intuitively, as the white rectangle show in the first row, Cal-SFDA achieves better result for the tail class `bike'. In comparison, both the source-only ECE and LD fail to predict the hub part. Our proposed method also shows better results for the object in depth, e.g., `car' and `sign', as shown in the second row of Figure \ref{fig:label-map}. These results demonstrate the effectiveness of our proposed method, especially for tiny objects.

\vspace{-1ex}
\section{Conclusion}\vspace{-1ex}
In summary, we introduce a novel methodology that effectively addresses the challenge of SFDA semantic segmentation with the aid of model calibration on both source and target sides. Our proposed model attains state-of-the-art results on two representative synthetic-to-real transfer tasks. Visualization of uncertainty and segmentation maps evidence that the recognition of tail classes can be improved. Our findings underscore the importance of ECE scores in SFDA semantic segmentation problems and provide promising strategies for the source and target model selections. 

\begin{acks}
This work was partially supported by Australian Research Council DE200101610.
\end{acks}
\newpage
\balance
\bibliographystyle{ACM-Reference-Format}
\bibliography{reference}


\begin{thebibliography}{58}


\ifx \showCODEN    \undefined \def \showCODEN     #1{\unskip}     \fi
\ifx \showDOI      \undefined \def \showDOI       #1{#1}\fi
\ifx \showISBNx    \undefined \def \showISBNx     #1{\unskip}     \fi
\ifx \showISBNxiii \undefined \def \showISBNxiii  #1{\unskip}     \fi
\ifx \showISSN     \undefined \def \showISSN      #1{\unskip}     \fi
\ifx \showLCCN     \undefined \def \showLCCN      #1{\unskip}     \fi
\ifx \shownote     \undefined \def \shownote      #1{#1}          \fi
\ifx \showarticletitle \undefined \def \showarticletitle #1{#1}   \fi
\ifx \showURL      \undefined \def \showURL       {\relax}        \fi
\providecommand\bibfield[2]{#2}
\providecommand\bibinfo[2]{#2}
\providecommand\natexlab[1]{#1}
\providecommand\showeprint[2][]{arXiv:#2}

\bibitem[Bohdal et~al\mbox{.}(2021)]%
        {DBLP:journals/corr/abs-2106-09613}
\bibfield{author}{\bibinfo{person}{Ondrej Bohdal}, \bibinfo{person}{Yongxin
  Yang}, {and} \bibinfo{person}{Timothy~M. Hospedales}.}
  \bibinfo{year}{2021}\natexlab{}.
\newblock \showarticletitle{Meta-Calibration: Meta-Learning of Model
  Calibration Using Differentiable Expected Calibration Error}.
\newblock \bibinfo{journal}{\emph{CoRR}}  \bibinfo{volume}{abs/2106.09613}
  (\bibinfo{year}{2021}).
\newblock
\showeprint[arXiv]{2106.09613}
\urldef\tempurl%
\url{https://arxiv.org/abs/2106.09613}
\showURL{%
\tempurl}


\bibitem[Chen et~al\mbox{.}(2018)]%
        {DBLP:journals/pami/ChenPKMY18}
\bibfield{author}{\bibinfo{person}{Liang{-}Chieh Chen}, \bibinfo{person}{George
  Papandreou}, \bibinfo{person}{Iasonas Kokkinos}, \bibinfo{person}{Kevin
  Murphy}, {and} \bibinfo{person}{Alan~L. Yuille}.}
  \bibinfo{year}{2018}\natexlab{}.
\newblock \showarticletitle{DeepLab: Semantic Image Segmentation with Deep
  Convolutional Nets, Atrous Convolution, and Fully Connected CRFs}.
\newblock \bibinfo{journal}{\emph{{IEEE} Trans. Pattern Anal. Mach. Intell.}}
  \bibinfo{volume}{40}, \bibinfo{number}{4} (\bibinfo{year}{2018}),
  \bibinfo{pages}{834--848}.
\newblock
\urldef\tempurl%
\url{https://doi.org/10.1109/TPAMI.2017.2699184}
\showDOI{\tempurl}


\bibitem[Chen et~al\mbox{.}(2021a)]%
        {DBLP:conf/iccv/ChenLQ0H0021}
\bibfield{author}{\bibinfo{person}{Zhi Chen}, \bibinfo{person}{Yadan Luo},
  \bibinfo{person}{Ruihong Qiu}, \bibinfo{person}{Sen Wang},
  \bibinfo{person}{Zi Huang}, \bibinfo{person}{Jingjing Li}, {and}
  \bibinfo{person}{Zheng Zhang}.} \bibinfo{year}{2021}\natexlab{a}.
\newblock \showarticletitle{Semantics Disentangling for Generalized Zero-Shot
  Learning}. In \bibinfo{booktitle}{\emph{2021 {IEEE/CVF} International
  Conference on Computer Vision, {ICCV} 2021, Montreal, QC, Canada, October
  10-17, 2021}}. \bibinfo{publisher}{{IEEE}}, \bibinfo{pages}{8692--8700}.
\newblock
\urldef\tempurl%
\url{https://doi.org/10.1109/ICCV48922.2021.00859}
\showDOI{\tempurl}


\bibitem[Chen et~al\mbox{.}(2021b)]%
        {DBLP:conf/mm/ChenL0QLH21}
\bibfield{author}{\bibinfo{person}{Zhi Chen}, \bibinfo{person}{Yadan Luo},
  \bibinfo{person}{Sen Wang}, \bibinfo{person}{Ruihong Qiu},
  \bibinfo{person}{Jingjing Li}, {and} \bibinfo{person}{Zi Huang}.}
  \bibinfo{year}{2021}\natexlab{b}.
\newblock \showarticletitle{Mitigating Generation Shifts for Generalized
  Zero-Shot Learning}. In \bibinfo{booktitle}{\emph{{MM} '21: {ACM} Multimedia
  Conference, Virtual Event, China, October 20 - 24, 2021}},
  \bibfield{editor}{\bibinfo{person}{Heng~Tao Shen}, \bibinfo{person}{Yueting
  Zhuang}, \bibinfo{person}{John~R. Smith}, \bibinfo{person}{Yang Yang},
  \bibinfo{person}{Pablo C{\'{e}}sar}, \bibinfo{person}{Florian Metze}, {and}
  \bibinfo{person}{Balakrishnan Prabhakaran}} (Eds.).
  \bibinfo{publisher}{{ACM}}, \bibinfo{pages}{844--852}.
\newblock
\urldef\tempurl%
\url{https://doi.org/10.1145/3474085.3475258}
\showDOI{\tempurl}


\bibitem[Chen et~al\mbox{.}(2020)]%
        {DBLP:conf/mm/Chen00H20}
\bibfield{author}{\bibinfo{person}{Zhi Chen}, \bibinfo{person}{Sen Wang},
  \bibinfo{person}{Jingjing Li}, {and} \bibinfo{person}{Zi Huang}.}
  \bibinfo{year}{2020}\natexlab{}.
\newblock \showarticletitle{Rethinking Generative Zero-Shot Learning: An
  Ensemble Learning Perspective for Recognising Visual Patches}. In
  \bibinfo{booktitle}{\emph{{MM} '20: The 28th {ACM} International Conference
  on Multimedia, Virtual Event / Seattle, WA, USA, October 12-16, 2020}},
  \bibfield{editor}{\bibinfo{person}{Chang~Wen Chen}, \bibinfo{person}{Rita
  Cucchiara}, \bibinfo{person}{Xian{-}Sheng Hua}, \bibinfo{person}{Guo{-}Jun
  Qi}, \bibinfo{person}{Elisa Ricci}, \bibinfo{person}{Zhengyou Zhang}, {and}
  \bibinfo{person}{Roger Zimmermann}} (Eds.). \bibinfo{publisher}{{ACM}},
  \bibinfo{pages}{3413--3421}.
\newblock
\urldef\tempurl%
\url{https://doi.org/10.1145/3394171.3413813}
\showDOI{\tempurl}


\bibitem[Cordts et~al\mbox{.}(2016)]%
        {DBLP:conf/cvpr/CordtsORREBFRS16}
\bibfield{author}{\bibinfo{person}{Marius Cordts}, \bibinfo{person}{Mohamed
  Omran}, \bibinfo{person}{Sebastian Ramos}, \bibinfo{person}{Timo Rehfeld},
  \bibinfo{person}{Markus Enzweiler}, \bibinfo{person}{Rodrigo Benenson},
  \bibinfo{person}{Uwe Franke}, \bibinfo{person}{Stefan Roth}, {and}
  \bibinfo{person}{Bernt Schiele}.} \bibinfo{year}{2016}\natexlab{}.
\newblock \showarticletitle{The Cityscapes Dataset for Semantic Urban Scene
  Understanding}. In \bibinfo{booktitle}{\emph{2016 {IEEE} Conference on
  Computer Vision and Pattern Recognition, {CVPR} 2016, Las Vegas, NV, USA,
  June 27-30, 2016}}. \bibinfo{publisher}{{IEEE} Computer Society},
  \bibinfo{pages}{3213--3223}.
\newblock
\urldef\tempurl%
\url{https://doi.org/10.1109/CVPR.2016.350}
\showDOI{\tempurl}


\bibitem[Cui et~al\mbox{.}(2021)]%
        {DBLP:journals/corr/abs-2107-06154}
\bibfield{author}{\bibinfo{person}{Shuhao Cui}, \bibinfo{person}{Shuhui Wang},
  \bibinfo{person}{Junbao Zhuo}, \bibinfo{person}{Liang Li},
  \bibinfo{person}{Qingming Huang}, {and} \bibinfo{person}{Qi Tian}.}
  \bibinfo{year}{2021}\natexlab{}.
\newblock \showarticletitle{Fast Batch Nuclear-norm Maximization and
  Minimization for Robust Domain Adaptation}.
\newblock \bibinfo{journal}{\emph{CoRR}}  \bibinfo{volume}{abs/2107.06154}
  (\bibinfo{year}{2021}).
\newblock
\showeprint[arXiv]{2107.06154}
\urldef\tempurl%
\url{https://arxiv.org/abs/2107.06154}
\showURL{%
\tempurl}


\bibitem[DeGroot and Fienberg(1983)]%
        {degroot1983comparison}
\bibfield{author}{\bibinfo{person}{Morris~H DeGroot} {and}
  \bibinfo{person}{Stephen~E Fienberg}.} \bibinfo{year}{1983}\natexlab{}.
\newblock \showarticletitle{The comparison and evaluation of forecasters}.
\newblock \bibinfo{journal}{\emph{Journal of the Royal Statistical Society:
  Series D (The Statistician)}} \bibinfo{volume}{32}, \bibinfo{number}{1-2}
  (\bibinfo{year}{1983}), \bibinfo{pages}{12--22}.
\newblock


\bibitem[Dong et~al\mbox{.}(2021a)]%
        {dong2021and}
\bibfield{author}{\bibinfo{person}{Jiahua Dong}, \bibinfo{person}{Yang Cong},
  \bibinfo{person}{Gan Sun}, \bibinfo{person}{Zhen Fang}, {and}
  \bibinfo{person}{Zhengming Ding}.} \bibinfo{year}{2021}\natexlab{a}.
\newblock \showarticletitle{Where and how to transfer: knowledge
  aggregation-induced transferability perception for unsupervised domain
  adaptation}.
\newblock \bibinfo{journal}{\emph{IEEE Transactions on Pattern Analysis and
  Machine Intelligence}} (\bibinfo{year}{2021}).
\newblock


\bibitem[Dong et~al\mbox{.}(2021b)]%
        {DBLP:conf/nips/DongFLSL21}
\bibfield{author}{\bibinfo{person}{Jiahua Dong}, \bibinfo{person}{Zhen Fang},
  \bibinfo{person}{Anjin Liu}, \bibinfo{person}{Gan Sun}, {and}
  \bibinfo{person}{Tongliang Liu}.} \bibinfo{year}{2021}\natexlab{b}.
\newblock \showarticletitle{Confident Anchor-Induced Multi-Source Free Domain
  Adaptation}. In \bibinfo{booktitle}{\emph{Advances in Neural Information
  Processing Systems 34: Annual Conference on Neural Information Processing
  Systems 2021, NeurIPS 2021, December 6-14, 2021, virtual}},
  \bibfield{editor}{\bibinfo{person}{Marc'Aurelio Ranzato},
  \bibinfo{person}{Alina Beygelzimer}, \bibinfo{person}{Yann~N. Dauphin},
  \bibinfo{person}{Percy Liang}, {and} \bibinfo{person}{Jennifer~Wortman
  Vaughan}} (Eds.). \bibinfo{pages}{2848--2860}.
\newblock
\urldef\tempurl%
\url{https://proceedings.neurips.cc/paper/2021/hash/168908dd3227b8358eababa07fcaf091-Abstract.html}
\showURL{%
\tempurl}


\bibitem[Guo et~al\mbox{.}(2017)]%
        {DBLP:conf/icml/GuoPSW17}
\bibfield{author}{\bibinfo{person}{Chuan Guo}, \bibinfo{person}{Geoff Pleiss},
  \bibinfo{person}{Yu Sun}, {and} \bibinfo{person}{Kilian~Q. Weinberger}.}
  \bibinfo{year}{2017}\natexlab{}.
\newblock \showarticletitle{On Calibration of Modern Neural Networks}. In
  \bibinfo{booktitle}{\emph{Proceedings of the 34th International Conference on
  Machine Learning, {ICML} 2017, Sydney, NSW, Australia, 6-11 August 2017}}
  \emph{(\bibinfo{series}{Proceedings of Machine Learning Research},
  Vol.~\bibinfo{volume}{70})}, \bibfield{editor}{\bibinfo{person}{Doina Precup}
  {and} \bibinfo{person}{Yee~Whye Teh}} (Eds.). \bibinfo{publisher}{{PMLR}},
  \bibinfo{pages}{1321--1330}.
\newblock
\urldef\tempurl%
\url{http://proceedings.mlr.press/v70/guo17a.html}
\showURL{%
\tempurl}


\bibitem[He et~al\mbox{.}(2016)]%
        {DBLP:conf/cvpr/HeZRS16}
\bibfield{author}{\bibinfo{person}{Kaiming He}, \bibinfo{person}{Xiangyu
  Zhang}, \bibinfo{person}{Shaoqing Ren}, {and} \bibinfo{person}{Jian Sun}.}
  \bibinfo{year}{2016}\natexlab{}.
\newblock \showarticletitle{Deep Residual Learning for Image Recognition}. In
  \bibinfo{booktitle}{\emph{2016 {IEEE} Conference on Computer Vision and
  Pattern Recognition, {CVPR} 2016, Las Vegas, NV, USA, June 27-30, 2016}}.
  \bibinfo{publisher}{{IEEE} Computer Society}, \bibinfo{pages}{770--778}.
\newblock
\urldef\tempurl%
\url{https://doi.org/10.1109/CVPR.2016.90}
\showDOI{\tempurl}


\bibitem[Huang et~al\mbox{.}(2021)]%
        {DBLP:conf/nips/HuangGXL21}
\bibfield{author}{\bibinfo{person}{Jiaxing Huang}, \bibinfo{person}{Dayan
  Guan}, \bibinfo{person}{Aoran Xiao}, {and} \bibinfo{person}{Shijian Lu}.}
  \bibinfo{year}{2021}\natexlab{}.
\newblock \showarticletitle{Model Adaptation: Historical Contrastive Learning
  for Unsupervised Domain Adaptation without Source Data}. In
  \bibinfo{booktitle}{\emph{Advances in Neural Information Processing Systems
  34: Annual Conference on Neural Information Processing Systems 2021, NeurIPS
  2021, December 6-14, 2021, virtual}},
  \bibfield{editor}{\bibinfo{person}{Marc'Aurelio Ranzato},
  \bibinfo{person}{Alina Beygelzimer}, \bibinfo{person}{Yann~N. Dauphin},
  \bibinfo{person}{Percy Liang}, {and} \bibinfo{person}{Jennifer~Wortman
  Vaughan}} (Eds.). \bibinfo{pages}{3635--3649}.
\newblock
\urldef\tempurl%
\url{https://proceedings.neurips.cc/paper/2021/hash/1dba5eed8838571e1c80af145184e515-Abstract.html}
\showURL{%
\tempurl}


\bibitem[Karandikar et~al\mbox{.}(2021)]%
        {DBLP:conf/nips/KarandikarCTLSM21}
\bibfield{author}{\bibinfo{person}{Archit Karandikar},
  \bibinfo{person}{Nicholas Cain}, \bibinfo{person}{Dustin Tran},
  \bibinfo{person}{Balaji Lakshminarayanan}, \bibinfo{person}{Jonathon Shlens},
  \bibinfo{person}{Michael~C. Mozer}, {and} \bibinfo{person}{Becca Roelofs}.}
  \bibinfo{year}{2021}\natexlab{}.
\newblock \showarticletitle{Soft Calibration Objectives for Neural Networks}.
  In \bibinfo{booktitle}{\emph{Advances in Neural Information Processing
  Systems 34: Annual Conference on Neural Information Processing Systems 2021,
  NeurIPS 2021, December 6-14, 2021, virtual}},
  \bibfield{editor}{\bibinfo{person}{Marc'Aurelio Ranzato},
  \bibinfo{person}{Alina Beygelzimer}, \bibinfo{person}{Yann~N. Dauphin},
  \bibinfo{person}{Percy Liang}, {and} \bibinfo{person}{Jennifer~Wortman
  Vaughan}} (Eds.). \bibinfo{pages}{29768--29779}.
\newblock
\urldef\tempurl%
\url{https://proceedings.neurips.cc/paper/2021/hash/f8905bd3df64ace64a68e154ba72f24c-Abstract.html}
\showURL{%
\tempurl}


\bibitem[Kim et~al\mbox{.}(2019)]%
        {DBLP:conf/iccv/KimYYK19}
\bibfield{author}{\bibinfo{person}{Youngdong Kim}, \bibinfo{person}{Junho Yim},
  \bibinfo{person}{Juseung Yun}, {and} \bibinfo{person}{Junmo Kim}.}
  \bibinfo{year}{2019}\natexlab{}.
\newblock \showarticletitle{{NLNL:} Negative Learning for Noisy Labels}. In
  \bibinfo{booktitle}{\emph{2019 {IEEE/CVF} International Conference on
  Computer Vision, {ICCV} 2019, Seoul, Korea (South), October 27 - November 2,
  2019}}. \bibinfo{publisher}{{IEEE}}, \bibinfo{pages}{101--110}.
\newblock
\urldef\tempurl%
\url{https://doi.org/10.1109/ICCV.2019.00019}
\showDOI{\tempurl}


\bibitem[Kouw and Loog(2021)]%
        {DBLP:journals/pami/KouwL21}
\bibfield{author}{\bibinfo{person}{Wouter~M. Kouw} {and} \bibinfo{person}{Marco
  Loog}.} \bibinfo{year}{2021}\natexlab{}.
\newblock \showarticletitle{A Review of Domain Adaptation without Target
  Labels}.
\newblock \bibinfo{journal}{\emph{{IEEE} Trans. Pattern Anal. Mach. Intell.}}
  \bibinfo{volume}{43}, \bibinfo{number}{3} (\bibinfo{year}{2021}),
  \bibinfo{pages}{766--785}.
\newblock
\urldef\tempurl%
\url{https://doi.org/10.1109/TPAMI.2019.2945942}
\showDOI{\tempurl}


\bibitem[Kull et~al\mbox{.}(2019)]%
        {DBLP:journals/corr/abs-1910-12656}
\bibfield{author}{\bibinfo{person}{Meelis Kull}, \bibinfo{person}{Miquel
  Perell{\'{o}}{-}Nieto}, \bibinfo{person}{Markus K{\"{a}}ngsepp},
  \bibinfo{person}{Telmo de~Menezes~e Silva~Filho}, \bibinfo{person}{Hao Song},
  {and} \bibinfo{person}{Peter~A. Flach}.} \bibinfo{year}{2019}\natexlab{}.
\newblock \showarticletitle{Beyond temperature scaling: Obtaining
  well-calibrated multiclass probabilities with Dirichlet calibration}.
\newblock \bibinfo{journal}{\emph{CoRR}}  \bibinfo{volume}{abs/1910.12656}
  (\bibinfo{year}{2019}).
\newblock
\showeprint[arXiv]{1910.12656}
\urldef\tempurl%
\url{http://arxiv.org/abs/1910.12656}
\showURL{%
\tempurl}


\bibitem[Kumar et~al\mbox{.}(2019)]%
        {DBLP:conf/nips/KumarLM19}
\bibfield{author}{\bibinfo{person}{Ananya Kumar}, \bibinfo{person}{Percy
  Liang}, {and} \bibinfo{person}{Tengyu Ma}.} \bibinfo{year}{2019}\natexlab{}.
\newblock \showarticletitle{Verified Uncertainty Calibration}. In
  \bibinfo{booktitle}{\emph{Advances in Neural Information Processing Systems
  32: Annual Conference on Neural Information Processing Systems 2019, NeurIPS
  2019, December 8-14, 2019, Vancouver, BC, Canada}},
  \bibfield{editor}{\bibinfo{person}{Hanna~M. Wallach}, \bibinfo{person}{Hugo
  Larochelle}, \bibinfo{person}{Alina Beygelzimer}, \bibinfo{person}{Florence
  d'Alch{\'{e}}{-}Buc}, \bibinfo{person}{Emily~B. Fox}, {and}
  \bibinfo{person}{Roman Garnett}} (Eds.). \bibinfo{pages}{3787--3798}.
\newblock
\urldef\tempurl%
\url{https://proceedings.neurips.cc/paper/2019/hash/f8c0c968632845cd133308b1a494967f-Abstract.html}
\showURL{%
\tempurl}


\bibitem[Kundu et~al\mbox{.}(2022)]%
        {DBLP:conf/icml/KunduKBMKJR22}
\bibfield{author}{\bibinfo{person}{Jogendra~Nath Kundu},
  \bibinfo{person}{Akshay~R. Kulkarni}, \bibinfo{person}{Suvaansh Bhambri},
  \bibinfo{person}{Deepesh Mehta}, \bibinfo{person}{Shreyas~Anand Kulkarni},
  \bibinfo{person}{Varun Jampani}, {and} \bibinfo{person}{Venkatesh~Babu
  Radhakrishnan}.} \bibinfo{year}{2022}\natexlab{}.
\newblock \showarticletitle{Balancing Discriminability and Transferability for
  Source-Free Domain Adaptation}. In \bibinfo{booktitle}{\emph{International
  Conference on Machine Learning, {ICML} 2022, 17-23 July 2022, Baltimore,
  Maryland, {USA}}} \emph{(\bibinfo{series}{Proceedings of Machine Learning
  Research}, Vol.~\bibinfo{volume}{162})},
  \bibfield{editor}{\bibinfo{person}{Kamalika Chaudhuri},
  \bibinfo{person}{Stefanie Jegelka}, \bibinfo{person}{Le~Song},
  \bibinfo{person}{Csaba Szepesv{\'{a}}ri}, \bibinfo{person}{Gang Niu}, {and}
  \bibinfo{person}{Sivan Sabato}} (Eds.). \bibinfo{publisher}{{PMLR}},
  \bibinfo{pages}{11710--11728}.
\newblock
\urldef\tempurl%
\url{https://proceedings.mlr.press/v162/kundu22a.html}
\showURL{%
\tempurl}


\bibitem[Kundu et~al\mbox{.}(2021)]%
        {DBLP:conf/iccv/KunduKSJB21}
\bibfield{author}{\bibinfo{person}{Jogendra~Nath Kundu},
  \bibinfo{person}{Akshay~R. Kulkarni}, \bibinfo{person}{Amit Singh},
  \bibinfo{person}{Varun Jampani}, {and} \bibinfo{person}{R.~Venkatesh Babu}.}
  \bibinfo{year}{2021}\natexlab{}.
\newblock \showarticletitle{Generalize then Adapt: Source-Free Domain Adaptive
  Semantic Segmentation}. In \bibinfo{booktitle}{\emph{2021 {IEEE/CVF}
  International Conference on Computer Vision, {ICCV} 2021, Montreal, QC,
  Canada, October 10-17, 2021}}. \bibinfo{publisher}{{IEEE}},
  \bibinfo{pages}{7026--7036}.
\newblock
\urldef\tempurl%
\url{https://doi.org/10.1109/ICCV48922.2021.00696}
\showDOI{\tempurl}


\bibitem[Li et~al\mbox{.}(2020)]%
        {DBLP:conf/eccv/LiKLWY20}
\bibfield{author}{\bibinfo{person}{Guangrui Li}, \bibinfo{person}{Guoliang
  Kang}, \bibinfo{person}{Wu Liu}, \bibinfo{person}{Yunchao Wei}, {and}
  \bibinfo{person}{Yi Yang}.} \bibinfo{year}{2020}\natexlab{}.
\newblock \showarticletitle{Content-Consistent Matching for Domain Adaptive
  Semantic Segmentation}. In \bibinfo{booktitle}{\emph{Computer Vision - {ECCV}
  2020 - 16th European Conference, Glasgow, UK, August 23-28, 2020,
  Proceedings, Part {XIV}}} \emph{(\bibinfo{series}{Lecture Notes in Computer
  Science}, Vol.~\bibinfo{volume}{12359})},
  \bibfield{editor}{\bibinfo{person}{Andrea Vedaldi}, \bibinfo{person}{Horst
  Bischof}, \bibinfo{person}{Thomas Brox}, {and} \bibinfo{person}{Jan{-}Michael
  Frahm}} (Eds.). \bibinfo{publisher}{Springer}, \bibinfo{pages}{440--456}.
\newblock
\urldef\tempurl%
\url{https://doi.org/10.1007/978-3-030-58568-6\_26}
\showDOI{\tempurl}


\bibitem[Li et~al\mbox{.}(2022)]%
        {DBLP:conf/cvpr/LiLHZJZ22}
\bibfield{author}{\bibinfo{person}{Ruihuang Li}, \bibinfo{person}{Shuai Li},
  \bibinfo{person}{Chenhang He}, \bibinfo{person}{Yabin Zhang},
  \bibinfo{person}{Xu Jia}, {and} \bibinfo{person}{Lei Zhang}.}
  \bibinfo{year}{2022}\natexlab{}.
\newblock \showarticletitle{Class-Balanced Pixel-Level Self-Labeling for Domain
  Adaptive Semantic Segmentation}. In \bibinfo{booktitle}{\emph{{IEEE/CVF}
  Conference on Computer Vision and Pattern Recognition, {CVPR} 2022, New
  Orleans, LA, USA, June 18-24, 2022}}. \bibinfo{publisher}{{IEEE}},
  \bibinfo{pages}{11583--11593}.
\newblock
\urldef\tempurl%
\url{https://doi.org/10.1109/CVPR52688.2022.01130}
\showDOI{\tempurl}


\bibitem[Liang et~al\mbox{.}(2020)]%
        {DBLP:conf/icml/LiangHF20}
\bibfield{author}{\bibinfo{person}{Jian Liang}, \bibinfo{person}{Dapeng Hu},
  {and} \bibinfo{person}{Jiashi Feng}.} \bibinfo{year}{2020}\natexlab{}.
\newblock \showarticletitle{Do We Really Need to Access the Source Data? Source
  Hypothesis Transfer for Unsupervised Domain Adaptation}. In
  \bibinfo{booktitle}{\emph{Proceedings of the 37th International Conference on
  Machine Learning, {ICML} 2020, 13-18 July 2020, Virtual Event}}
  \emph{(\bibinfo{series}{Proceedings of Machine Learning Research},
  Vol.~\bibinfo{volume}{119})}. \bibinfo{publisher}{{PMLR}},
  \bibinfo{pages}{6028--6039}.
\newblock
\urldef\tempurl%
\url{http://proceedings.mlr.press/v119/liang20a.html}
\showURL{%
\tempurl}


\bibitem[Linardos et~al\mbox{.}(2021)]%
        {DBLP:conf/iccv/LinardosKPB21}
\bibfield{author}{\bibinfo{person}{Akis Linardos}, \bibinfo{person}{Matthias
  K{\"{u}}mmerer}, \bibinfo{person}{Ori Press}, {and} \bibinfo{person}{Matthias
  Bethge}.} \bibinfo{year}{2021}\natexlab{}.
\newblock \showarticletitle{DeepGaze {IIE:} Calibrated prediction in and
  out-of-domain for state-of-the-art saliency modeling}. In
  \bibinfo{booktitle}{\emph{2021 {IEEE/CVF} International Conference on
  Computer Vision, {ICCV} 2021, Montreal, QC, Canada, October 10-17, 2021}}.
  \bibinfo{publisher}{{IEEE}}, \bibinfo{pages}{12899--12908}.
\newblock
\urldef\tempurl%
\url{https://doi.org/10.1109/ICCV48922.2021.01268}
\showDOI{\tempurl}


\bibitem[Liu et~al\mbox{.}(2021)]%
        {DBLP:conf/cvpr/LiuZW21}
\bibfield{author}{\bibinfo{person}{Yuang Liu}, \bibinfo{person}{Wei Zhang},
  {and} \bibinfo{person}{Jun Wang}.} \bibinfo{year}{2021}\natexlab{}.
\newblock \showarticletitle{Source-Free Domain Adaptation for Semantic
  Segmentation}. In \bibinfo{booktitle}{\emph{{IEEE} Conference on Computer
  Vision and Pattern Recognition, {CVPR} 2021, virtual, June 19-25, 2021}}.
  \bibinfo{publisher}{Computer Vision Foundation / {IEEE}},
  \bibinfo{pages}{1215--1224}.
\newblock
\urldef\tempurl%
\url{https://doi.org/10.1109/CVPR46437.2021.00127}
\showDOI{\tempurl}


\bibitem[Luo et~al\mbox{.}(2019)]%
        {luo2019significance}
\bibfield{author}{\bibinfo{person}{Yawei Luo}, \bibinfo{person}{Ping Liu},
  \bibinfo{person}{Tao Guan}, \bibinfo{person}{Junqing Yu}, {and}
  \bibinfo{person}{Yi Yang}.} \bibinfo{year}{2019}\natexlab{}.
\newblock \showarticletitle{Significance-aware information bottleneck for
  domain adaptive semantic segmentation}. In
  \bibinfo{booktitle}{\emph{Proceedings of the IEEE/CVF International
  Conference on Computer Vision}}. \bibinfo{pages}{6778--6787}.
\newblock


\bibitem[Luo et~al\mbox{.}(2021)]%
        {luo2021category}
\bibfield{author}{\bibinfo{person}{Yawei Luo}, \bibinfo{person}{Ping Liu},
  \bibinfo{person}{Liang Zheng}, \bibinfo{person}{Tao Guan},
  \bibinfo{person}{Junqing Yu}, {and} \bibinfo{person}{Yi Yang}.}
  \bibinfo{year}{2021}\natexlab{}.
\newblock \showarticletitle{Category-level adversarial adaptation for semantic
  segmentation using purified features}.
\newblock \bibinfo{journal}{\emph{IEEE Transactions on Pattern Analysis and
  Machine Intelligence}} (\bibinfo{year}{2021}).
\newblock


\bibitem[Minderer et~al\mbox{.}(2021)]%
        {DBLP:conf/nips/MindererDRHZHTL21}
\bibfield{author}{\bibinfo{person}{Matthias Minderer}, \bibinfo{person}{Josip
  Djolonga}, \bibinfo{person}{Rob Romijnders}, \bibinfo{person}{Frances Hubis},
  \bibinfo{person}{Xiaohua Zhai}, \bibinfo{person}{Neil Houlsby},
  \bibinfo{person}{Dustin Tran}, {and} \bibinfo{person}{Mario Lucic}.}
  \bibinfo{year}{2021}\natexlab{}.
\newblock \showarticletitle{Revisiting the Calibration of Modern Neural
  Networks}. In \bibinfo{booktitle}{\emph{Advances in Neural Information
  Processing Systems 34: Annual Conference on Neural Information Processing
  Systems 2021, NeurIPS 2021, December 6-14, 2021, virtual}},
  \bibfield{editor}{\bibinfo{person}{Marc'Aurelio Ranzato},
  \bibinfo{person}{Alina Beygelzimer}, \bibinfo{person}{Yann~N. Dauphin},
  \bibinfo{person}{Percy Liang}, {and} \bibinfo{person}{Jennifer~Wortman
  Vaughan}} (Eds.). \bibinfo{pages}{15682--15694}.
\newblock
\urldef\tempurl%
\url{https://proceedings.neurips.cc/paper/2021/hash/8420d359404024567b5aefda1231af24-Abstract.html}
\showURL{%
\tempurl}


\bibitem[Mukhoti et~al\mbox{.}(2020)]%
        {DBLP:conf/nips/MukhotiKSGTD20}
\bibfield{author}{\bibinfo{person}{Jishnu Mukhoti}, \bibinfo{person}{Viveka
  Kulharia}, \bibinfo{person}{Amartya Sanyal}, \bibinfo{person}{Stuart
  Golodetz}, \bibinfo{person}{Philip H.~S. Torr}, {and}
  \bibinfo{person}{Puneet~K. Dokania}.} \bibinfo{year}{2020}\natexlab{}.
\newblock \showarticletitle{Calibrating Deep Neural Networks using Focal Loss}.
  In \bibinfo{booktitle}{\emph{Advances in Neural Information Processing
  Systems 33: Annual Conference on Neural Information Processing Systems 2020,
  NeurIPS 2020, December 6-12, 2020, virtual}},
  \bibfield{editor}{\bibinfo{person}{Hugo Larochelle},
  \bibinfo{person}{Marc'Aurelio Ranzato}, \bibinfo{person}{Raia Hadsell},
  \bibinfo{person}{Maria{-}Florina Balcan}, {and} \bibinfo{person}{Hsuan{-}Tien
  Lin}} (Eds.).
\newblock
\urldef\tempurl%
\url{https://proceedings.neurips.cc/paper/2020/hash/aeb7b30ef1d024a76f21a1d40e30c302-Abstract.html}
\showURL{%
\tempurl}


\bibitem[Naeini et~al\mbox{.}(2015)]%
        {DBLP:conf/aaai/NaeiniCH15}
\bibfield{author}{\bibinfo{person}{Mahdi~Pakdaman Naeini},
  \bibinfo{person}{Gregory~F. Cooper}, {and} \bibinfo{person}{Milos
  Hauskrecht}.} \bibinfo{year}{2015}\natexlab{}.
\newblock \showarticletitle{Obtaining Well Calibrated Probabilities Using
  Bayesian Binning}. In \bibinfo{booktitle}{\emph{Proceedings of the
  Twenty-Ninth {AAAI} Conference on Artificial Intelligence, January 25-30,
  2015, Austin, Texas, {USA}}}, \bibfield{editor}{\bibinfo{person}{Blai Bonet}
  {and} \bibinfo{person}{Sven Koenig}} (Eds.). \bibinfo{publisher}{{AAAI}
  Press}, \bibinfo{pages}{2901--2907}.
\newblock
\urldef\tempurl%
\url{http://www.aaai.org/ocs/index.php/AAAI/AAAI15/paper/view/9667}
\showURL{%
\tempurl}


\bibitem[Pan et~al\mbox{.}(2020)]%
        {DBLP:conf/cvpr/PanSRLK20}
\bibfield{author}{\bibinfo{person}{Fei Pan}, \bibinfo{person}{Inkyu Shin},
  \bibinfo{person}{Fran{\c{c}}ois Rameau}, \bibinfo{person}{Seokju Lee}, {and}
  \bibinfo{person}{In~So Kweon}.} \bibinfo{year}{2020}\natexlab{}.
\newblock \showarticletitle{Unsupervised Intra-Domain Adaptation for Semantic
  Segmentation Through Self-Supervision}. In \bibinfo{booktitle}{\emph{2020
  {IEEE/CVF} Conference on Computer Vision and Pattern Recognition, {CVPR}
  2020, Seattle, WA, USA, June 13-19, 2020}}. \bibinfo{publisher}{Computer
  Vision Foundation / {IEEE}}, \bibinfo{pages}{3763--3772}.
\newblock
\urldef\tempurl%
\url{https://doi.org/10.1109/CVPR42600.2020.00382}
\showDOI{\tempurl}


\bibitem[Park et~al\mbox{.}(2020)]%
        {DBLP:conf/aistats/ParkBWL20}
\bibfield{author}{\bibinfo{person}{Sangdon Park}, \bibinfo{person}{Osbert
  Bastani}, \bibinfo{person}{James Weimer}, {and} \bibinfo{person}{Insup Lee}.}
  \bibinfo{year}{2020}\natexlab{}.
\newblock \showarticletitle{Calibrated Prediction with Covariate Shift via
  Unsupervised Domain Adaptation}. In \bibinfo{booktitle}{\emph{The 23rd
  International Conference on Artificial Intelligence and Statistics, {AISTATS}
  2020, 26-28 August 2020, Online [Palermo, Sicily, Italy]}}
  \emph{(\bibinfo{series}{Proceedings of Machine Learning Research},
  Vol.~\bibinfo{volume}{108})}, \bibfield{editor}{\bibinfo{person}{Silvia
  Chiappa} {and} \bibinfo{person}{Roberto Calandra}} (Eds.).
  \bibinfo{publisher}{{PMLR}}, \bibinfo{pages}{3219--3229}.
\newblock
\urldef\tempurl%
\url{http://proceedings.mlr.press/v108/park20b.html}
\showURL{%
\tempurl}


\bibitem[Platt et~al\mbox{.}(1999)]%
        {platt1999probabilistic}
\bibfield{author}{\bibinfo{person}{John Platt} {et~al\mbox{.}}}
  \bibinfo{year}{1999}\natexlab{}.
\newblock \showarticletitle{Probabilistic outputs for support vector machines
  and comparisons to regularized likelihood methods}.
\newblock \bibinfo{journal}{\emph{Advances in large margin classifiers}}
  \bibinfo{volume}{10}, \bibinfo{number}{3} (\bibinfo{year}{1999}),
  \bibinfo{pages}{61--74}.
\newblock


\bibitem[Prabhu et~al\mbox{.}(2021)]%
        {DBLP:journals/corr/abs-2107-10140}
\bibfield{author}{\bibinfo{person}{Viraj Prabhu}, \bibinfo{person}{Shivam
  Khare}, \bibinfo{person}{Deeksha Kartik}, {and} \bibinfo{person}{Judy
  Hoffman}.} \bibinfo{year}{2021}\natexlab{}.
\newblock \showarticletitle{{S4T:} Source-free domain adaptation for semantic
  segmentation via self-supervised selective self-training}.
\newblock \bibinfo{journal}{\emph{CoRR}}  \bibinfo{volume}{abs/2107.10140}
  (\bibinfo{year}{2021}).
\newblock
\showeprint[arXiv]{2107.10140}
\urldef\tempurl%
\url{https://arxiv.org/abs/2107.10140}
\showURL{%
\tempurl}


\bibitem[Prechelt(2012)]%
        {DBLP:series/lncs/Prechelt12}
\bibfield{author}{\bibinfo{person}{Lutz Prechelt}.}
  \bibinfo{year}{2012}\natexlab{}.
\newblock \showarticletitle{Early Stopping - But When?}
\newblock In \bibinfo{booktitle}{\emph{Neural Networks: Tricks of the Trade -
  Second Edition}}, \bibfield{editor}{\bibinfo{person}{Gr{\'{e}}goire
  Montavon}, \bibinfo{person}{Genevieve~B. Orr}, {and}
  \bibinfo{person}{Klaus{-}Robert M{\"{u}}ller}} (Eds.).
  \bibinfo{series}{Lecture Notes in Computer Science},
  Vol.~\bibinfo{volume}{7700}. \bibinfo{publisher}{Springer},
  \bibinfo{pages}{53--67}.
\newblock
\urldef\tempurl%
\url{https://doi.org/10.1007/978-3-642-35289-8\_5}
\showDOI{\tempurl}


\bibitem[Richter et~al\mbox{.}(2016)]%
        {DBLP:conf/eccv/RichterVRK16}
\bibfield{author}{\bibinfo{person}{Stephan~R. Richter}, \bibinfo{person}{Vibhav
  Vineet}, \bibinfo{person}{Stefan Roth}, {and} \bibinfo{person}{Vladlen
  Koltun}.} \bibinfo{year}{2016}\natexlab{}.
\newblock \showarticletitle{Playing for Data: Ground Truth from Computer
  Games}. In \bibinfo{booktitle}{\emph{Computer Vision - {ECCV} 2016 - 14th
  European Conference, Amsterdam, The Netherlands, October 11-14, 2016,
  Proceedings, Part {II}}} \emph{(\bibinfo{series}{Lecture Notes in Computer
  Science}, Vol.~\bibinfo{volume}{9906})},
  \bibfield{editor}{\bibinfo{person}{Bastian Leibe}, \bibinfo{person}{Jiri
  Matas}, \bibinfo{person}{Nicu Sebe}, {and} \bibinfo{person}{Max Welling}}
  (Eds.). \bibinfo{publisher}{Springer}, \bibinfo{pages}{102--118}.
\newblock
\urldef\tempurl%
\url{https://doi.org/10.1007/978-3-319-46475-6\_7}
\showDOI{\tempurl}


\bibitem[Ros et~al\mbox{.}(2016)]%
        {DBLP:conf/cvpr/RosSMVL16}
\bibfield{author}{\bibinfo{person}{Germ{\'{a}}n Ros}, \bibinfo{person}{Laura
  Sellart}, \bibinfo{person}{Joanna Materzynska}, \bibinfo{person}{David
  V{\'{a}}zquez}, {and} \bibinfo{person}{Antonio~M. L{\'{o}}pez}.}
  \bibinfo{year}{2016}\natexlab{}.
\newblock \showarticletitle{The {SYNTHIA} Dataset: {A} Large Collection of
  Synthetic Images for Semantic Segmentation of Urban Scenes}. In
  \bibinfo{booktitle}{\emph{2016 {IEEE} Conference on Computer Vision and
  Pattern Recognition, {CVPR} 2016, Las Vegas, NV, USA, June 27-30, 2016}}.
  \bibinfo{publisher}{{IEEE} Computer Society}, \bibinfo{pages}{3234--3243}.
\newblock
\urldef\tempurl%
\url{https://doi.org/10.1109/CVPR.2016.352}
\showDOI{\tempurl}


\bibitem[Tsai et~al\mbox{.}(2018)]%
        {DBLP:conf/cvpr/TsaiHSS0C18}
\bibfield{author}{\bibinfo{person}{Yi{-}Hsuan Tsai},
  \bibinfo{person}{Wei{-}Chih Hung}, \bibinfo{person}{Samuel Schulter},
  \bibinfo{person}{Kihyuk Sohn}, \bibinfo{person}{Ming{-}Hsuan Yang}, {and}
  \bibinfo{person}{Manmohan Chandraker}.} \bibinfo{year}{2018}\natexlab{}.
\newblock \showarticletitle{Learning to Adapt Structured Output Space for
  Semantic Segmentation}. In \bibinfo{booktitle}{\emph{2018 {IEEE} Conference
  on Computer Vision and Pattern Recognition, {CVPR} 2018, Salt Lake City, UT,
  USA, June 18-22, 2018}}. \bibinfo{publisher}{Computer Vision Foundation /
  {IEEE} Computer Society}, \bibinfo{pages}{7472--7481}.
\newblock
\urldef\tempurl%
\url{https://doi.org/10.1109/CVPR.2018.00780}
\showDOI{\tempurl}


\bibitem[Tsai et~al\mbox{.}(2019)]%
        {DBLP:conf/iccv/TsaiSSC19}
\bibfield{author}{\bibinfo{person}{Yi{-}Hsuan Tsai}, \bibinfo{person}{Kihyuk
  Sohn}, \bibinfo{person}{Samuel Schulter}, {and} \bibinfo{person}{Manmohan
  Chandraker}.} \bibinfo{year}{2019}\natexlab{}.
\newblock \showarticletitle{Domain Adaptation for Structured Output via
  Discriminative Patch Representations}. In \bibinfo{booktitle}{\emph{2019
  {IEEE/CVF} International Conference on Computer Vision, {ICCV} 2019, Seoul,
  Korea (South), October 27 - November 2, 2019}}. \bibinfo{publisher}{{IEEE}},
  \bibinfo{pages}{1456--1465}.
\newblock
\urldef\tempurl%
\url{https://doi.org/10.1109/ICCV.2019.00154}
\showDOI{\tempurl}


\bibitem[Vu et~al\mbox{.}(2019)]%
        {DBLP:conf/cvpr/VuJBCP19}
\bibfield{author}{\bibinfo{person}{Tuan{-}Hung Vu}, \bibinfo{person}{Himalaya
  Jain}, \bibinfo{person}{Maxime Bucher}, \bibinfo{person}{Matthieu Cord},
  {and} \bibinfo{person}{Patrick P{\'{e}}rez}.}
  \bibinfo{year}{2019}\natexlab{}.
\newblock \showarticletitle{{ADVENT:} Adversarial Entropy Minimization for
  Domain Adaptation in Semantic Segmentation}. In
  \bibinfo{booktitle}{\emph{{IEEE} Conference on Computer Vision and Pattern
  Recognition, {CVPR} 2019, Long Beach, CA, USA, June 16-20, 2019}}.
  \bibinfo{publisher}{Computer Vision Foundation / {IEEE}},
  \bibinfo{pages}{2517--2526}.
\newblock
\urldef\tempurl%
\url{https://doi.org/10.1109/CVPR.2019.00262}
\showDOI{\tempurl}


\bibitem[Wald et~al\mbox{.}(2021)]%
        {DBLP:conf/nips/WaldFGS21}
\bibfield{author}{\bibinfo{person}{Yoav Wald}, \bibinfo{person}{Amir Feder},
  \bibinfo{person}{Daniel Greenfeld}, {and} \bibinfo{person}{Uri Shalit}.}
  \bibinfo{year}{2021}\natexlab{}.
\newblock \showarticletitle{On Calibration and Out-of-Domain Generalization}.
  In \bibinfo{booktitle}{\emph{Advances in Neural Information Processing
  Systems 34: Annual Conference on Neural Information Processing Systems 2021,
  NeurIPS 2021, December 6-14, 2021, virtual}},
  \bibfield{editor}{\bibinfo{person}{Marc'Aurelio Ranzato},
  \bibinfo{person}{Alina Beygelzimer}, \bibinfo{person}{Yann~N. Dauphin},
  \bibinfo{person}{Percy Liang}, {and} \bibinfo{person}{Jennifer~Wortman
  Vaughan}} (Eds.). \bibinfo{pages}{2215--2227}.
\newblock
\urldef\tempurl%
\url{https://proceedings.neurips.cc/paper/2021/hash/118bd558033a1016fcc82560c65cca5f-Abstract.html}
\showURL{%
\tempurl}


\bibitem[Wang et~al\mbox{.}(2021)]%
        {DBLP:conf/iclr/WangSLOD21}
\bibfield{author}{\bibinfo{person}{Dequan Wang}, \bibinfo{person}{Evan
  Shelhamer}, \bibinfo{person}{Shaoteng Liu}, \bibinfo{person}{Bruno~A.
  Olshausen}, {and} \bibinfo{person}{Trevor Darrell}.}
  \bibinfo{year}{2021}\natexlab{}.
\newblock \showarticletitle{Tent: Fully Test-Time Adaptation by Entropy
  Minimization}. In \bibinfo{booktitle}{\emph{9th International Conference on
  Learning Representations, {ICLR} 2021, Virtual Event, Austria, May 3-7,
  2021}}. \bibinfo{publisher}{OpenReview.net}.
\newblock
\urldef\tempurl%
\url{https://openreview.net/forum?id=uXl3bZLkr3c}
\showURL{%
\tempurl}


\bibitem[Wang et~al\mbox{.}(2018)]%
        {DBLP:conf/mm/WangFCYHY18}
\bibfield{author}{\bibinfo{person}{Jindong Wang}, \bibinfo{person}{Wenjie
  Feng}, \bibinfo{person}{Yiqiang Chen}, \bibinfo{person}{Han Yu},
  \bibinfo{person}{Meiyu Huang}, {and} \bibinfo{person}{Philip~S. Yu}.}
  \bibinfo{year}{2018}\natexlab{}.
\newblock \showarticletitle{Visual Domain Adaptation with Manifold Embedded
  Distribution Alignment}. In \bibinfo{booktitle}{\emph{2018 {ACM} Multimedia
  Conference on Multimedia Conference, {MM} 2018, Seoul, Republic of Korea,
  October 22-26, 2018}}, \bibfield{editor}{\bibinfo{person}{Susanne Boll},
  \bibinfo{person}{Kyoung~Mu Lee}, \bibinfo{person}{Jiebo Luo},
  \bibinfo{person}{Wenwu Zhu}, \bibinfo{person}{Hyeran Byun},
  \bibinfo{person}{Chang~Wen Chen}, \bibinfo{person}{Rainer Lienhart}, {and}
  \bibinfo{person}{Tao Mei}} (Eds.). \bibinfo{publisher}{{ACM}},
  \bibinfo{pages}{402--410}.
\newblock
\urldef\tempurl%
\url{https://doi.org/10.1145/3240508.3240512}
\showDOI{\tempurl}


\bibitem[Wang and Deng(2018)]%
        {DBLP:journals/ijon/WangD18}
\bibfield{author}{\bibinfo{person}{Mei Wang} {and} \bibinfo{person}{Weihong
  Deng}.} \bibinfo{year}{2018}\natexlab{}.
\newblock \showarticletitle{Deep visual domain adaptation: {A} survey}.
\newblock \bibinfo{journal}{\emph{Neurocomputing}}  \bibinfo{volume}{312}
  (\bibinfo{year}{2018}), \bibinfo{pages}{135--153}.
\newblock
\urldef\tempurl%
\url{https://doi.org/10.1016/j.neucom.2018.05.083}
\showDOI{\tempurl}


\bibitem[Wang et~al\mbox{.}(2019)]%
        {DBLP:conf/iccv/0001MCLY019}
\bibfield{author}{\bibinfo{person}{Yisen Wang}, \bibinfo{person}{Xingjun Ma},
  \bibinfo{person}{Zaiyi Chen}, \bibinfo{person}{Yuan Luo},
  \bibinfo{person}{Jinfeng Yi}, {and} \bibinfo{person}{James Bailey}.}
  \bibinfo{year}{2019}\natexlab{}.
\newblock \showarticletitle{Symmetric Cross Entropy for Robust Learning With
  Noisy Labels}. In \bibinfo{booktitle}{\emph{2019 {IEEE/CVF} International
  Conference on Computer Vision, {ICCV} 2019, Seoul, Korea (South), October 27
  - November 2, 2019}}. \bibinfo{publisher}{{IEEE}}, \bibinfo{pages}{322--330}.
\newblock
\urldef\tempurl%
\url{https://doi.org/10.1109/ICCV.2019.00041}
\showDOI{\tempurl}


\bibitem[Wang et~al\mbox{.}(2022)]%
        {DBLP:conf/icmcs/WangLZ0H22}
\bibfield{author}{\bibinfo{person}{Zixin Wang}, \bibinfo{person}{Yadan Luo},
  \bibinfo{person}{Peng{-}Fei Zhang}, \bibinfo{person}{Sen Wang}, {and}
  \bibinfo{person}{Zi Huang}.} \bibinfo{year}{2022}\natexlab{}.
\newblock \showarticletitle{Discovering Domain Disentanglement for Generalized
  Multi-Source Domain Adaptation}. In \bibinfo{booktitle}{\emph{{IEEE}
  International Conference on Multimedia and Expo, {ICME} 2022, Taipei, Taiwan,
  July 18-22, 2022}}. \bibinfo{publisher}{{IEEE}}, \bibinfo{pages}{1--6}.
\newblock
\urldef\tempurl%
\url{https://doi.org/10.1109/ICME52920.2022.9859733}
\showDOI{\tempurl}


\bibitem[Xie et~al\mbox{.}(2023)]%
        {DBLP:journals/corr/abs-2302-13824}
\bibfield{author}{\bibinfo{person}{Mixue Xie}, \bibinfo{person}{Shuang Li},
  \bibinfo{person}{Rui Zhang}, {and} \bibinfo{person}{Chi~Harold Liu}.}
  \bibinfo{year}{2023}\natexlab{}.
\newblock \showarticletitle{Dirichlet-based Uncertainty Calibration for Active
  Domain Adaptation}.
\newblock \bibinfo{journal}{\emph{CoRR}}  \bibinfo{volume}{abs/2302.13824}
  (\bibinfo{year}{2023}).
\newblock
\urldef\tempurl%
\url{https://doi.org/10.48550/arXiv.2302.13824}
\showDOI{\tempurl}
\showeprint[arXiv]{2302.13824}


\bibitem[Yang et~al\mbox{.}(2022)]%
        {DBLP:conf/icmcs/YangKH22}
\bibfield{author}{\bibinfo{person}{Cheng{-}Yu Yang},
  \bibinfo{person}{Yuan{-}Jhe Kuo}, {and} \bibinfo{person}{Chiou{-}Ting Hsu}.}
  \bibinfo{year}{2022}\natexlab{}.
\newblock \showarticletitle{Source Free Domain Adaptation for Semantic
  Segmentation via Distribution Transfer and Adaptive Class-Balanced
  Self-Training}. In \bibinfo{booktitle}{\emph{{IEEE} International Conference
  on Multimedia and Expo, {ICME} 2022, Taipei, Taiwan, July 18-22, 2022}}.
  \bibinfo{publisher}{{IEEE}}, \bibinfo{pages}{1--6}.
\newblock
\urldef\tempurl%
\url{https://doi.org/10.1109/ICME52920.2022.9859581}
\showDOI{\tempurl}


\bibitem[Yang and Soatto(2020)]%
        {DBLP:conf/cvpr/0001S20}
\bibfield{author}{\bibinfo{person}{Yanchao Yang} {and} \bibinfo{person}{Stefano
  Soatto}.} \bibinfo{year}{2020}\natexlab{}.
\newblock \showarticletitle{{FDA:} Fourier Domain Adaptation for Semantic
  Segmentation}. In \bibinfo{booktitle}{\emph{2020 {IEEE/CVF} Conference on
  Computer Vision and Pattern Recognition, {CVPR} 2020, Seattle, WA, USA, June
  13-19, 2020}}. \bibinfo{publisher}{Computer Vision Foundation / {IEEE}},
  \bibinfo{pages}{4084--4094}.
\newblock
\urldef\tempurl%
\url{https://doi.org/10.1109/CVPR42600.2020.00414}
\showDOI{\tempurl}


\bibitem[Ye et~al\mbox{.}(2021)]%
        {DBLP:conf/mm/Ye0OY21}
\bibfield{author}{\bibinfo{person}{Mucong Ye}, \bibinfo{person}{Jing Zhang},
  \bibinfo{person}{Jinpeng Ouyang}, {and} \bibinfo{person}{Ding Yuan}.}
  \bibinfo{year}{2021}\natexlab{}.
\newblock \showarticletitle{Source Data-free Unsupervised Domain Adaptation for
  Semantic Segmentation}. In \bibinfo{booktitle}{\emph{{MM} '21: {ACM}
  Multimedia Conference, Virtual Event, China, October 20 - 24, 2021}},
  \bibfield{editor}{\bibinfo{person}{Heng~Tao Shen}, \bibinfo{person}{Yueting
  Zhuang}, \bibinfo{person}{John~R. Smith}, \bibinfo{person}{Yang Yang},
  \bibinfo{person}{Pablo Cesar}, \bibinfo{person}{Florian Metze}, {and}
  \bibinfo{person}{Balakrishnan Prabhakaran}} (Eds.).
  \bibinfo{publisher}{{ACM}}, \bibinfo{pages}{2233--2242}.
\newblock
\urldef\tempurl%
\url{https://doi.org/10.1145/3474085.3475384}
\showDOI{\tempurl}


\bibitem[You et~al\mbox{.}(2021)]%
        {DBLP:conf/mm/YouLZCH21}
\bibfield{author}{\bibinfo{person}{Fuming You}, \bibinfo{person}{Jingjing Li},
  \bibinfo{person}{Lei Zhu}, \bibinfo{person}{Zhi Chen}, {and}
  \bibinfo{person}{Zi Huang}.} \bibinfo{year}{2021}\natexlab{}.
\newblock \showarticletitle{Domain Adaptive Semantic Segmentation without
  Source Data}. In \bibinfo{booktitle}{\emph{{MM} '21: {ACM} Multimedia
  Conference, Virtual Event, China, October 20 - 24, 2021}},
  \bibfield{editor}{\bibinfo{person}{Heng~Tao Shen}, \bibinfo{person}{Yueting
  Zhuang}, \bibinfo{person}{John~R. Smith}, \bibinfo{person}{Yang Yang},
  \bibinfo{person}{Pablo Cesar}, \bibinfo{person}{Florian Metze}, {and}
  \bibinfo{person}{Balakrishnan Prabhakaran}} (Eds.).
  \bibinfo{publisher}{{ACM}}, \bibinfo{pages}{3293--3302}.
\newblock
\urldef\tempurl%
\url{https://doi.org/10.1145/3474085.3475482}
\showDOI{\tempurl}


\bibitem[Yu et~al\mbox{.}(2021)]%
        {DBLP:conf/aaai/YuZDHDZ21}
\bibfield{author}{\bibinfo{person}{Fei Yu}, \bibinfo{person}{Mo Zhang},
  \bibinfo{person}{Hexin Dong}, \bibinfo{person}{Sheng Hu},
  \bibinfo{person}{Bin Dong}, {and} \bibinfo{person}{Li Zhang}.}
  \bibinfo{year}{2021}\natexlab{}.
\newblock \showarticletitle{{DAST:} Unsupervised Domain Adaptation in Semantic
  Segmentation Based on Discriminator Attention and Self-Training}. In
  \bibinfo{booktitle}{\emph{Thirty-Fifth {AAAI} Conference on Artificial
  Intelligence, {AAAI} 2021, Thirty-Third Conference on Innovative Applications
  of Artificial Intelligence, {IAAI} 2021, The Eleventh Symposium on
  Educational Advances in Artificial Intelligence, {EAAI} 2021, Virtual Event,
  February 2-9, 2021}}. \bibinfo{publisher}{{AAAI} Press},
  \bibinfo{pages}{10754--10762}.
\newblock
\urldef\tempurl%
\url{https://ojs.aaai.org/index.php/AAAI/article/view/17285}
\showURL{%
\tempurl}


\bibitem[Zadrozny and Elkan(2001)]%
        {DBLP:conf/icml/ZadroznyE01}
\bibfield{author}{\bibinfo{person}{Bianca Zadrozny} {and}
  \bibinfo{person}{Charles Elkan}.} \bibinfo{year}{2001}\natexlab{}.
\newblock \showarticletitle{Obtaining calibrated probability estimates from
  decision trees and naive Bayesian classifiers}. In
  \bibinfo{booktitle}{\emph{Proceedings of the Eighteenth International
  Conference on Machine Learning {(ICML} 2001), Williams College, Williamstown,
  MA, USA, June 28 - July 1, 2001}},
  \bibfield{editor}{\bibinfo{person}{Carla~E. Brodley} {and}
  \bibinfo{person}{Andrea~Pohoreckyj Danyluk}} (Eds.).
  \bibinfo{publisher}{Morgan Kaufmann}, \bibinfo{pages}{609--616}.
\newblock


\bibitem[Zhang et~al\mbox{.}(2021)]%
        {DBLP:journals/corr/abs-2110-04596}
\bibfield{author}{\bibinfo{person}{Yifan Zhang}, \bibinfo{person}{Bingyi Kang},
  \bibinfo{person}{Bryan Hooi}, \bibinfo{person}{Shuicheng Yan}, {and}
  \bibinfo{person}{Jiashi Feng}.} \bibinfo{year}{2021}\natexlab{}.
\newblock \showarticletitle{Deep Long-Tailed Learning: {A} Survey}.
\newblock \bibinfo{journal}{\emph{CoRR}}  \bibinfo{volume}{abs/2110.04596}
  (\bibinfo{year}{2021}).
\newblock
\showeprint[arXiv]{2110.04596}
\urldef\tempurl%
\url{https://arxiv.org/abs/2110.04596}
\showURL{%
\tempurl}


\bibitem[Zhao et~al\mbox{.}(2022)]%
        {DBLP:journals/tcsv/ZhaoZLLS22}
\bibfield{author}{\bibinfo{person}{Yuyang Zhao}, \bibinfo{person}{Zhun Zhong},
  \bibinfo{person}{Zhiming Luo}, \bibinfo{person}{Gim~Hee Lee}, {and}
  \bibinfo{person}{Nicu Sebe}.} \bibinfo{year}{2022}\natexlab{}.
\newblock \showarticletitle{Source-Free Open Compound Domain Adaptation in
  Semantic Segmentation}.
\newblock \bibinfo{journal}{\emph{{IEEE} Trans. Circuits Syst. Video Technol.}}
  \bibinfo{volume}{32}, \bibinfo{number}{10} (\bibinfo{year}{2022}),
  \bibinfo{pages}{7019--7032}.
\newblock
\urldef\tempurl%
\url{https://doi.org/10.1109/TCSVT.2022.3179021}
\showDOI{\tempurl}


\bibitem[Zou et~al\mbox{.}(2018)]%
        {DBLP:conf/eccv/ZouYKW18}
\bibfield{author}{\bibinfo{person}{Yang Zou}, \bibinfo{person}{Zhiding Yu},
  \bibinfo{person}{B.~V. K.~Vijaya Kumar}, {and} \bibinfo{person}{Jinsong
  Wang}.} \bibinfo{year}{2018}\natexlab{}.
\newblock \showarticletitle{Unsupervised Domain Adaptation for Semantic
  Segmentation via Class-Balanced Self-training}. In
  \bibinfo{booktitle}{\emph{Computer Vision - {ECCV} 2018 - 15th European
  Conference, Munich, Germany, September 8-14, 2018, Proceedings, Part {III}}}
  \emph{(\bibinfo{series}{Lecture Notes in Computer Science},
  Vol.~\bibinfo{volume}{11207})}, \bibfield{editor}{\bibinfo{person}{Vittorio
  Ferrari}, \bibinfo{person}{Martial Hebert}, \bibinfo{person}{Cristian
  Sminchisescu}, {and} \bibinfo{person}{Yair Weiss}} (Eds.).
  \bibinfo{publisher}{Springer}, \bibinfo{pages}{297--313}.
\newblock
\urldef\tempurl%
\url{https://doi.org/10.1007/978-3-030-01219-9\_18}
\showDOI{\tempurl}


\bibitem[Zou et~al\mbox{.}(2019)]%
        {DBLP:conf/iccv/ZouYLKW19}
\bibfield{author}{\bibinfo{person}{Yang Zou}, \bibinfo{person}{Zhiding Yu},
  \bibinfo{person}{Xiaofeng Liu}, \bibinfo{person}{B.~V. K.~Vijaya Kumar},
  {and} \bibinfo{person}{Jinsong Wang}.} \bibinfo{year}{2019}\natexlab{}.
\newblock \showarticletitle{Confidence Regularized Self-Training}. In
  \bibinfo{booktitle}{\emph{2019 {IEEE/CVF} International Conference on
  Computer Vision, {ICCV} 2019, Seoul, Korea (South), October 27 - November 2,
  2019}}. \bibinfo{publisher}{{IEEE}}, \bibinfo{pages}{5981--5990}.
\newblock
\urldef\tempurl%
\url{https://doi.org/10.1109/ICCV.2019.00608}
\showDOI{\tempurl}


\bibitem[Zou et~al\mbox{.}(2021)]%
        {DBLP:conf/iclr/ZouZZLBHP21}
\bibfield{author}{\bibinfo{person}{Yuliang Zou}, \bibinfo{person}{Zizhao
  Zhang}, \bibinfo{person}{Han Zhang}, \bibinfo{person}{Chun{-}Liang Li},
  \bibinfo{person}{Xiao Bian}, \bibinfo{person}{Jia{-}Bin Huang}, {and}
  \bibinfo{person}{Tomas Pfister}.} \bibinfo{year}{2021}\natexlab{}.
\newblock \showarticletitle{PseudoSeg: Designing Pseudo Labels for Semantic
  Segmentation}. In \bibinfo{booktitle}{\emph{9th International Conference on
  Learning Representations, {ICLR} 2021, Virtual Event, Austria, May 3-7,
  2021}}. \bibinfo{publisher}{OpenReview.net}.
\newblock
\urldef\tempurl%
\url{https://openreview.net/forum?id=-TwO99rbVRu}
\showURL{%
\tempurl}


\end{thebibliography}

\newpage
\appendix
\title{\Huge Appendix}
\begin{algorithm}[h]
\SetKwInOut{Input}{Input}
\SetKwInOut{Output}{Output}

\Input{Training dataset $\mathcal{D}_S$, validation dataset $\mathcal{D}_S^v$;\
Segmentation model $f$ with parameters $\theta$;\
Batch size $B$;\
Number of epochs $E$;\
}
\Output{Best source checkpoint $\theta^*$.}
\BlankLine
Initialize pre-trained source model\;
\BlankLine
\tcp{Source model training}
\For{$i=0$ \KwTo $E$}{
\For{$j=0$ \KwTo $|\mathcal{D}_S|/B$}{
Sample a batch of images and their corresponding labels from $\mathcal{D}_S$\;
Pass this batch of samples through segmentation model $f$\;
Compute the loss in Eq (1)\;
Compute the gradients of the loss with respect to the model parameters\;
Update the parameters by SGD\;
}
Save the checkpoint $\theta_i$ into the checkpoint pool $\Theta$
}
\BlankLine
\tcp{Source model selection}
Randomly sample a subset $\mathcal{D}_S^v$ of $\mathcal{D}_S$\;
\For{$\theta_i$ in $\Theta$}{
  Calculate $\text{mean}(\text{ECE}_{\theta_i})$ $\text{max}(\text{ECE}_{\theta_i})$ and $\text{min}(\text{ECE}_{\theta_i})$ for $\mathcal{D}_S^v$\;
}
\tcp{Choose the best checkpoint}
Find out the source checkpoint $\theta^*$ by Eq (6). \
\caption{Source model training and model selection}
\label{alg:src}
\end{algorithm}
\section{Algorithm} \label{sec:alg}

The Algorithm \ref{alg:src} depicts the training process of the source model, which entails the fine-tuning of the pre-trained model by $\mathcal{L}\text{seg}$ on the source data while simultaneously optimizing the calibration (ECE) loss $\mathcal{L}_{\text{ECE}_\text{diff}}$. Upon completion of the training, the proposed model selection strategy is introduced to determine the optimal source checkpoint $f(\cdot;\theta^*)$.
\begin{algorithm}[b]
\SetKwInOut{Input}{Input}
\SetKwInOut{Output}{Output}
\Input{Source domain: $\mathcal{D}_S$; Source validation set $\mathcal{D}_S^v$;
Segmentation model $f$ with parameter $\theta^*$;
Value net $v$ with parameter $\phi$ ;}
\Output{Trained value net}
\BlankLine
\tcp{Freeze the segmentation network}
\For{$i=0$ \KwTo $E$}{
\For{$j=0$ \KwTo $|\mathcal{D}_S|/B$}{
Pass the source data batch $x_j$ through $f$ up to layer $l$ to obtain the feature maps $f_l(x_j)$ \;
\tcp{The ECE is predicted per image}
Feed $f_l(x_j)$ to the value net $v$ to obtain the predicted ECE score $\widehat{\text{ECE}}_j$\;
Calculate the true ECE score from the frozen segmentation network: $\text{ECE}_j$ \;
Compute the loss $\mathcal{L}_\text{match}$ in Eq (7) to minimize the difference between $\text{ECE}_j$ and $\widehat{\text{ECE}}_j$\;
Compute the gradients of the loss with respect to the value net parameters\;
Update the parameter $\phi$ by SGD;
}
Stop once $\mathcal{L}_\text{match}$ converged \;
}
Find the best checkpoint as $\phi^*$ with the smallest $\mathcal{L}_\text{match}$ for the source set $\mathcal{D}_S^v$.
\caption{Value net training}
\label{alg:valnet}
\end{algorithm}

After determining the best source checkpoint $\theta^*$, keep it frozen. The segmentation feature extractor is then linked to an additional branch called value net $v$. This supplementary branch, as presented in Algorithm \ref{alg:valnet}, is created to predict the ECE score of the given source data per image, striving to attain a score as similar as possible to the actual ECE score.

\begin{algorithm}[h]
\SetKwInOut{Input}{Input}
\SetKwInOut{Output}{Output}

\Input{Unlabeled target data $\mathcal{D}_T$, pre-trained segmentation model $f$ with parameter $\theta^*$, value net $v$ with parameter $\phi^*$; Label ratio: $\delta$}
\Output{The resulting segmentation model $f$ with parameter $\theta'$}
\BlankLine
\tcp{ECE-guided thresholding}
\For{$x_j$ in $\mathcal{D}_T$}{
Pass $x_j$ through $f$ to obtain predicted probability maps\;
Calculate the modified confidence as Eq (8) presents\;
For the predicted confidence of class $c$, save it into the class confidence set $\mathfrak{R}^c$ in descending order\;
}
Compute class-wise confidence threshold in Eq (10)\;
\BlankLine
\tcp{Generate pseudo labels}
\For{$x_i$ in $\mathcal{D}_T$}{
Follow the confidence threshold to select global pseudo labels\;
Select a certain proportion of pixels with the predicted result per image as local pseudo labels\;
Fuse the global and local pseudo-labels together\
}
Get $\hat{\mathcal{Y}}_T^{\prime}$ as the target pseudo-label set\;
\BlankLine
\tcp{Fine-tune model on the target data}
\For{$i=0$ \KwTo $E$}{
Initialize variable $\text{Ent}_i=0$ to store the entropy of each epoch\;
\For{$j=0$ \KwTo $|\mathcal{D}_T|/B$}{
Sample a batch of images and their corresponding pseudo labels from $\mathcal{D}_T$\;
Compute the loss in Eq (18), including $\mathcal{L}_{\text {sce }}$, $\mathcal{L}_\text{ent}$, and $\mathcal{L}_{\text {neg }}$\;
$\text{Ent}_i \mathrel{+}= \mathcal{L}_\text{ent}$\;
Compute the gradients of the loss with respect to the model parameters\;
Update the parameters of layers (exclude feature extractor and value net) by SGD \;
}
$\text{Ent}_i = \text{Ent}_i/|\mathcal{D}_T|$\;
}
\BlankLine
\tcp{Target checkpoint selection}
Choose the one with the smallest $\text{Ent}_i$ as the best target checkpoint $\theta'$.
\caption{Calibration-aware self-training}
\label{alg:tar}
\end{algorithm}

We first use estimated ECE from the value net to guide the thresholding and pseudo-labeling process during target adaptation in Algorithm \ref{alg:tar}. After assigning pseudo-labels to the target data, a statistic warm-up module is then conducted to stabilize the adaptation. Finally, weighted self-training is applied to mitigate the domain shift between the source and the target.

\section{Method} \label{sec:method}
\subsection{Details of Statistic Warm-up}
With the pseudo-labels acquired from the class-balanced thresholding, we can conduct supervised learning on the pseudo-labeled data within the target domain. While a naive adaptation approach would involve fine-tuning the entire network $\theta$, we observe that a large domain gap can lead to instability during the training process. We adopt a statistic warm-up strategy to mitigate this issue, following \cite{DBLP:conf/iclr/WangSLOD21}. Apart from updating the mean and variance by the statistics of the current batch, we also update the transformation parameters in the backward pass. Specifically, we update the transformation parameters $\gamma$ and $\beta$ of all BatchNorm layers by utilizing a supervised loss $\mathcal{L}_\text{tar}$:
\begin{equation} \tag{19}
    \gamma \leftarrow \gamma+\partial \mathcal{L}_\text{tar} / \partial \gamma, 
    \beta \leftarrow \beta+\partial \mathcal{L}_\text{tar} / \partial \beta.
\end{equation}
\subsection{Symmetric Cross Entropy Loss}
Despite being guided by the estimated ECE, the target adaptation process appears unstable due to the noisy and imbalanced pseudo-label set. To mitigate this issue, we employ symmetric cross-entropy (sce) loss \cite{DBLP:conf/iccv/0001MCLY019} instead of the conventional cross-entropy (CE) loss. The sce loss comprises CE loss and a reversed CE. Its purpose is not to enhance performance but to guide the adaptation process while preventing extreme performance outcomes, which will further benefit the target model selection process.

\begin{figure*}
\centering
\includegraphics[width=1\linewidth]{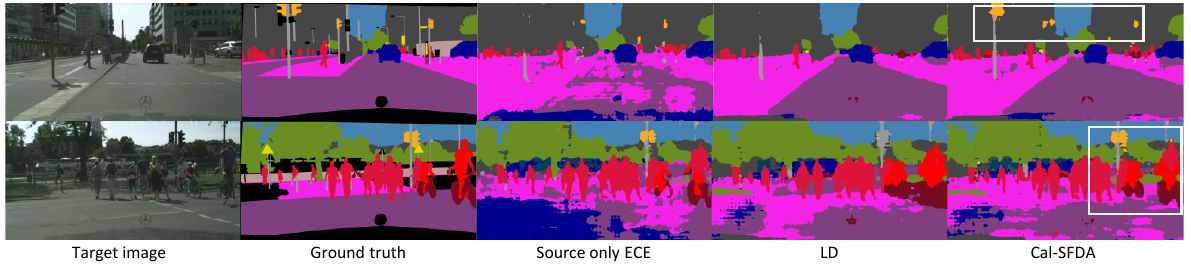}
\caption{More qualitative results of Cal-SFDA on Synthia→Cityscapes. White boxes are key areas.}\label{fig:label-map-viz}
\end{figure*}
\section{Experiment Details} \label{sec:exp}
\subsection{Datasets}
\textbf{GTA5} dataset \cite{DBLP:conf/eccv/RichterVRK16} consists of 24,966 synthesized frames captured from a video game. These are provided with pixel-level semantic annotations of 33 different classes. To facilitate the comparison with previous works, we focus on the 19 classes with the Cityscapes dataset for our experiments. \textbf{Synthia}. \cite{DBLP:conf/cvpr/RosSMVL16} Similarly, we employ the SYNTHIA-RAND-CITYSCAPES set, which consists of 9,400 synthesized images for training our models. During training, we utilize 16 shared common classes in SYNTHIA and Cityscapes. To evaluate our models' performance, we follow the protocol used in previous models and compare their performance on both 16-class and 13-class subsets of the Cityscapes dataset. \textbf{Cityscapes} \cite{DBLP:conf/cvpr/CordtsORREBFRS16} is a real-world dataset designed for autonomous driving scenarios, and it comprises images from 50 cities worldwide. The dataset includes 2975 images for training and 500 images for validation. Under the SFSS setting, the source-only model undergoes adapting on the unlabeled training set and is then evaluated on the validation set with full annotations.

To assess the ability of our proposed model to adapt to the target domain, we randomly select a small amount of the source data $\mathcal{D}_S^v$ as mentioned in Algorithm \ref{alg:src}, for the source model selection.

\subsection{Implementation details. }
We apply data augmentations that are identical to those used in \cite{DBLP:conf/mm/YouLZCH21}, which involve resizing the images and randomly cropping them to a size of $600 \times 600$. Furthermore, we perform horizontal flips and randomly scale the images between 0.5 and 1.5. The evaluation metric used is the mean intersection-over-union (mIoU), a commonly used metric in semantic segmentation tasks. During source pre-training, we first optimize the model by the pure cross-entropy loss, and then the differentiable ECE will join the training process, which could prevent the classification task from being dominated. In target adaptation, we train $6$ rounds for GTA5$\rightarrow$Cityscapes benchmarks and $1$ round for Synthia$\rightarrow$Cityscapes. 
To ensure a fair selection strategy, we selected the source checkpoint based on the minimum min/max/mean Expected Calibration Error on the randomly selected Synthia dataset $\mathcal{D}_S^v$. In the case of GTA5, we selected the checkpoint based on the minimum mean ECE.

\section{Qualitative Study}\label{sec:viz}
Figure \ref{fig:label-map-viz} shows the prediction map for the source-only ECE model, LD, and our proposed Cal-SFDA. The first row of the figure reveals that both the source-only ECE model and the proposed Cal-SFDA have a commendable performance in recognizing smaller objects, such as traffic lights. In between, Cal-SFDA could further fine-tune the object shape. Similar outcomes are evident in the second row of the figure, where our proposed model surpasses LD in terms of identifying traffic lights. Besides, other small objects like pedestrians and bicycles are also shown to be better than LD, as exhibited in comparison with the ground-truth label map, which suggests the effectiveness of our model.
\end{sloppypar}

\end{document}